\documentclass[11pt]{article}

\usepackage[final]{acl}

\usepackage{times}
\usepackage{latexsym}

\usepackage[T1]{fontenc}

\usepackage[utf8]{inputenc}

\usepackage{microtype}

\usepackage{inconsolata}

\usepackage{graphicx}
\usepackage{xspace}
\usepackage{algorithm}
\usepackage{algpseudocode}
\usepackage{amsmath}
\usepackage{multirow}
\usepackage{makecell}
\usepackage{enumitem}
\usepackage{wrapfig}
\usepackage{xspace}
\usepackage{booktabs}

\usepackage{caption}
\usepackage{hhline}

\usepackage{url}

\newcommand{\method}{PTCG\xspace}
\title{\method: Persona-guided Tree-based Counterargument Generation}

\author{
Eunbeen Son\textsuperscript{1} \hspace{.7cm} 
Yohan Jo\textsuperscript{2} \hspace{.7cm} 
Joonsuk Park\textsuperscript{3,4\textdagger} \hspace{.7cm} 
JinYeong Bak\textsuperscript{1\textdagger} \\
  \textsuperscript{1}Sungkyunkwan University \hspace{.4cm} 
  \textsuperscript{2}Seoul National University\\
  \textsuperscript{3}University of Richmond \hspace{.4cm} 
  \textsuperscript{4}NAVER Cloud\\
  \texttt{nabin111@g.skku.edu}, \texttt{yohan.jo@snu.ac.kr},\\
  \texttt{park@joonsuk.org}, \texttt{jy.bak@skku.edu} \\
  }

\begin{document}
\maketitle
\begingroup\def\thefootnote{\textdagger}\footnotetext{Corresponding authors}\endgroup
\renewcommand{\thefootnote}{\arabic{footnote}}
\begin{abstract}
  The ability to generate counterarguments is important for critical thinking and balanced discourse, yet existing approaches typically produce only a single counterargument, failing to capture the diversity and persuasiveness required in real-world debates.
To address this limitation, we propose Persona-guided Tree-based Counterargument Generation (\method), a framework that combines Tree-of-Thoughts–inspired stepwise generation and pruning with speaker persona selection.
By estimating the author’s persona from the original argument and incorporating speaker personas representing distinct perspectives, \method operationalizes perspective-taking and enables the generation of diverse counterarguments.
Results from LLM-as-a-Judge, classifier-based assessment, and human evaluations indicate that \method shows consistent improvements in both the diversity and persuasiveness of counterarguments compared to baseline methods.
\footnote{Code repo: \url{https://github.com/nabin0111/PTCG}}
\end{abstract}

\section{Introduction}
\label{sec:introduction}


The ability to generate effective counterarguments is a growing area of interest in computational argumentation~\citep{wang2023argument}.
A counterargument is not merely an expression of disagreement but a reasoned response that challenges an argument, exposing assumptions, logical gaps, or alternative perspectives.
Engaging with counterarguments is widely recognized as a key mechanism for fostering critical thinking, as it encourages individuals to evaluate evidence, weigh competing viewpoints, and refine reasoning~\citep{ennis2015critical, dekker2020teaching}.
The ability to generate effective counterarguments is especially important in contexts such as political debates, legal reasoning, and online discussions, where opposing views promote balanced discourse and better-informed decision-making~\citep{li-etal-2020-exploring-role, behrendt2025natural, zhang2025mitigating, gray2025generating}.

Existing counterargument generation approaches suffer from two critical limitations: 
First, most methods generate a counterargument focusing on a single point, failing to capture diverse perspectives on the subject.
This limitation stems from their reliance on a single strategy, such as attacking a weak premise~\citep{alshomary-etal-2021-counter}, directly refuting the conclusion~\citep{alshomary-wachsmuth-2023-conclusion}, or pointing out logical flaws~\citep{lin-etal-2023-argue}.
Second, counterarguments produced by existing approaches, including those generated with large language models (LLMs), often lack persuasiveness.
While \citet{chen-etal-2024-exploring-potential} demonstrate the potential of LLMs in argument generation, subsequent studies (e.g. \citealt{lu2025mind}, \citealt{plenz-etal-2025-argumentation}) reveal that generated arguments often lack value-based reasoning and scenario-driven perspectives essential for persuasiveness in human argumentation.
In addition, the aforementioned issue of narrow perspective also negatively impacts the overall persuasiveness of the counterargument.
This suggests that persuasiveness across diverse audiences depends on presenting arguments from multiple perspectives, making the resolution of this limitation a central challenge for counterargument generation.



To address these issues, we propose Persona-guided Tree-based Counterargument Generation (\method), a framework grounded in the theory of perspective-taking.
Perspective-taking, a social psychology concept of adopting other's standpoints during argumentation, foster empathy, reduce bias, and encourage reasoning from different viewpoints~\citep{batson1997perspective, green2000role}.
\method operationalize perspective-taking by guiding counterargument generation with pre-defined personas.
Specifically, after estimating the original author's persona from their argument, the framework selects personas from both similar and contrasting predefined persona clusters and uses them to guide counterargument generation.
The predefined persona clusters were created to organize about 50,000 personas, reducing redundancy and enabling efficient selection of diverse perspectives.
This design enables models to move beyond default stances and generate counterarguments that reflect a wider range of perspectives.
In addition, \method incorporates a Tree-of-Thoughts (ToT)–inspired reasoning procedure~\citep{yao2023tree}.
Multiple candidate reasoning paths 
called \emph{plans} 
for generating counterarguments are first generated, evaluated, and pruned.
Only the most promising plans are then expanded into full counterarguments.
This iterative process of generation and selection allows for diverse and persuasive reasoning paths.
By combining persona conditioning with a stepwise reasoning process inspired by Tree-of-Thoughts (ToT)~\citep{yao2023tree}, the framework iteratively generates and selects candidate plans and counterarguments, ultimately producing multiple counterarguments that capture both diversity and persuasiveness.

To evaluate \method, we conduct experiments with multiple LLMs using the discussion threads from the ChangeMyView subreddit, which cover a diverse range of real-world topics.
We combine LLM-as-a-Judge to assess the diversity and persuasiveness, general and targeted, as well as the stance and quality of the generated counterarguments.
We further incorporate classifier-based metrics for the key dimension of persuasiveness, providing a more comprehensive evaluation.
In addition, we conduct human evaluation, which not only complements the LLM-as-a-Judge results but also demonstrates persuasiveness across a diverse pool of evaluators, providing further evidence of applicability to the real audience.
Across these evaluations, \method consistently outperforms the baselines, producing counterarguments that are more diverse, persuasive, and higher in overall quality.

\section{Related Work}
\label{sec:related_work}
\subsection{Argument Generation}
\label{sec:argument_generation}
Early work on argument generation framed the task as a largely symbolic or rule-driven process.
\citet{sato-etal-2015-end} proposed a pipeline-based debating system, but such rule-based approaches were often brittle and lacked scalability.
To improve flexibility, \citet{wachsmuth-etal-2018-argumentation}  employed hand-crafted rhetorical patterns, while \citet{hua-wang-2018-neural} introduced neural argument generation augmented with retrieved evidence to ground arguments in factual content.
Together, these approaches laid important groundwork for argument and counterargument generation.

\subsection{Counterargument Generation}
\label{sec:counterargument_generation}
Recent studies on counterargument generation have primarily focused on exploiting explicit argument structures or predefined strategies.
For instance, \citet{alshomary-etal-2021-counter} generate counterarguments by attacking weak premises, while \citet{alshomary-wachsmuth-2023-conclusion} guide generation by modeling the conclusion of the original post.
\citet{lin-etal-2023-argue} instead operate at the sentence level, producing concise counterarguments for individual statements.
However, these approaches typically rely on a single strategy and generate only one counterargument, limiting their ability to explore diverse perspectives.
Unlike prior work, our framework aims to produce multiple distinct counterarguments grounded in diverse personas and reasoning trajectories.
To achieve this, we ground reasoning in diverse personas and explore alternative argumentative paths through tree-based stepwise generation.

While \citet{hu-etal-2025-debate} recently proposed a multi-agent framework in which personas debate to synthesize a single essay, our objective is fundamentally different.
Rather than merging multiple voices into a unified output, our approach allows each persona to independently develop its own reasoning.
As a result, diversity across personas is not merely an intermediate mechanism for producing a single response, but a primary objective of the framework itself.
By systematizing persona selection based on distance, our approach generates multiple distinct and persuasive counterarguments that reflect a broader spectrum of perspectives.

\subsection{Perspective-Taking}
\label{sec:perspective-taking}
Psychological studies highlight the persuasive power of perspective-taking and narrative immersion.
\citet{batson1997perspective} show that imagining how others feel fosters empathy and altruistic motivation.
\citet{green2000role} and \citet{mar2008function} suggest that narrative `transportation' enables simulated experience, which can influence attitudes more deeply than factual exposition.
More recently, \citet{bullock2021narratives} argue that narratives are persuasive partly because they are processed more fluently than non-narrative formats.
These findings motivate our use of perspective-taking-based generation to induce perspectival engagement and simulate meaningful disagreement.
Moreover, perspective-taking enables the incorporation of diverse viewpoints, making it possible to generate multiple counterarguments for a single post.

\subsection{Diverse-Audience Persuasion}
\label{sec:personalized_persuasion}
\citet{lukin-etal-2017-argument} show that persuasiveness depends on audience traits, motivating the use of personality-based analysis in argumentation.
Building on this, studies have explored personalization in persuasive dialogue: \citet{wang-etal-2019-persuasion} adapt strategies based on user traits, and \citet{al-khatib-etal-2020-exploiting} incorporate debaters' characteristics to improve persuasiveness prediction.
Recent work further demonstrates that LLMs can generate more persuasive messages when tailored to psychological profiles~\citep{matz2024potential}, modulate linguistic features according to personality cues~\citep{mieleszczenko2024dark}, and role-play personas to enhance empathy and strategy distribution~\citep{yang2025psyplay}.
Building on this line of research, we integrate audience diversity into persona-guided generation to produce both diverse and persuasive counterarguments.

\section{Persona-guided Tree-based Counterargument Generation}
\label{sec:method}
\begin{figure*}[!ht]
\centering
\includegraphics[width=0.95\textwidth]{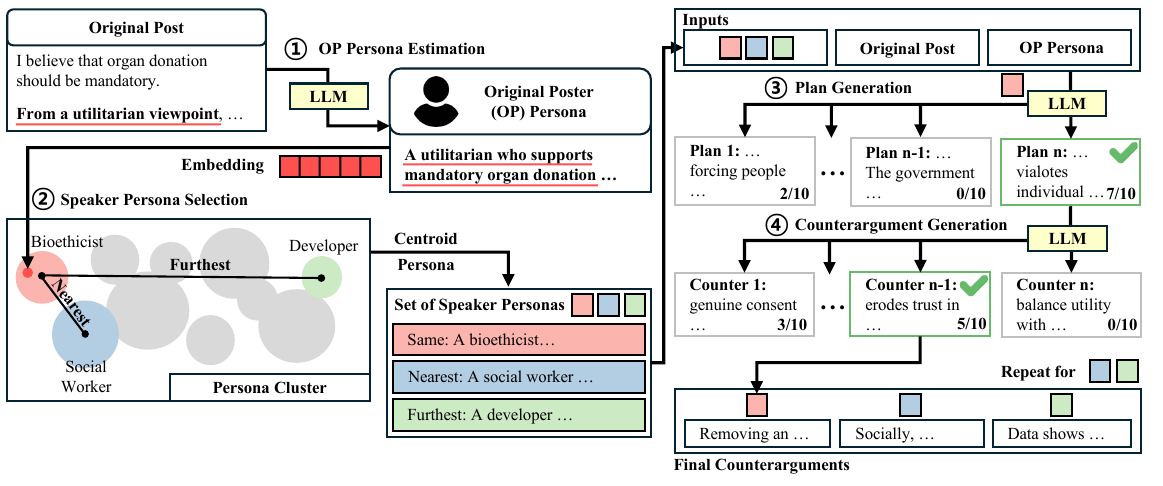}
\caption{Persona-guided Tree-based Counterargument Generation (\method). 
An LLM first extracts the author’s persona and conditions on distinct personas to produce multiple candidate reasoning plans. 
These plans are evaluated and pruned in a tree-based stepwise manner, and the most promising ones are expanded into three final counterarguments, each grounded in a different persona.}
\label{fig:method_description}
\end{figure*}





\subsection{Task Description}
\label{sec:task_description}
We define the \textbf{multiple distinct counterargument generation} to evaluate generating persuasive counterarguments that cover diverse perspectives on a given argument.
Specifically, the input is an 
\emph{argument}
consisting of a claim and one or more premises supporting it, and the output is a set of \emph{counterarguments}, each presenting a distinct perspective that challenges the original argument.
We detail the evaluation criteria in Section~\ref{sec:evaluation_metrics}.


\subsection{Perspective Taking: Persona Selection}
\label{sec:persona_selection}
To realize perspective-taking in counterargument generation, we draw on the notion of personas as proxies for diverse viewpoints.
Prior work has shown that incorporating personas into dialogue models allows them to go beyond generic language modeling, producing responses that are more consistent, human-like, and reflective of underlying experiences or viewpoints~\citep{zhang-etal-2018-personalizing, moon-etal-2024-virtual}.
We adopt the PersonaHub dataset~\citep{ge2024scaling}, a large-scale repository of about 50,000 personas.
Since using all personas is redundant and costly, we cluster similar ones into groups.
Each persona is embedded using OpenAI’s text-embedding-3-large (\citealt{OpenAI2024-embedding}; 3,072-dimensional), reduced to a 50-dimensional representation via UMAP~\citep{mcinnes2018umap}, and clustered into 39 groups\footnote{As shown in Appendix Table~\ref{tab:clustering_results_table}, the configuration with 39 clusters was selected as it achieves the highest Silhouette Score~\citep{rousseeuw1987silhouettes} while also maintaining a strong Calinski--Harabasz Index~\citep{calinski1974dendrite}, indicating the best balance between cohesion and separation.} using HDBSCAN~\citep{mcinnes2017hdbscan}.
Subsequently, we leverage inter-cluster distances to select counterargument speakers from distinct groups, ensuring that the generated counterarguments embody genuinely different perspectives.
To help readers better understand the semantic coherence and diversity encoded in these clusters, we provide representative persona examples from several clusters in Section~\ref{sec:cluster-examples}.
These examples illustrate how grouping similar personas enables systematic selection of viewpoints that truly differ in background, expertise, and worldview—an essential property for perspective-taking in counterargument generation.




\subsection{Persona-guided Tree-based Counterargument Generation (\method) Framework}
\label{sec:method_framework}
Our method, Persona-guided Tree-based Counterargument Generation (\method; Figure~\ref{fig:method_description} and Algorithm~\ref{algorithm:tree_of_personas}), integrates the Original Poster (OP) persona estimation, clustering-based speaker persona selection, and a Tree-of-Thoughts (ToT)~\citep{yao2023tree}-inspired stepwise generation process (Plan and Counterargument Generation and Selection).


\paragraph{Step 1: Original Poster (OP) Persona Estimation}
Given an original post, we estimate the OP persona using an LLM-based estimation prompt.
Here, the OP refers to the author of the original post.
This estimated persona represents the OP's beliefs, values, and worldview, and conditions subsequent counterargument generation.
The detailed prompt is provided in Appendix Figure~\ref{fig:op_estimation_prompt}.

\paragraph{Step 2: Speaker Persona Selection}
To balance diversity with interpretability, we set the number of personas to three, supported by observations from the CMV dataset where posts with multiple delta-awarded comments, indicating diverse, high-quality counterarguments, rarely exceeded three (see Appendix Figure~\ref{fig:number_of_delta_awarded_comments}).
We select personas based on their cluster distance from the OP: one from the \emph{same} cluster, one from the \emph{nearest} cluster, and one from the \emph{furthest} cluster.
For each cluster, no persona exactly matched the centroid; therefore, we used the closest in embedding space.
This setup ensures that the three chosen personas collectively reflect perspectives ranging from highly aligned to markedly divergent, thereby systematically probing how cluster distance influences counterargument generation and persuasiveness.

\paragraph{Step 3: Plan Generation and Selection}
Once the OP persona and speaker personas are determined, the generation process proceeds in a tree-based stepwise manner inspired by ToT~\citep{yao2023tree}.
For each persona, the LLM generates three candidate plans, each outlining a persuasive strategy for counterargument generation.
A voting procedure then evaluates these candidates in terms of whether they effectively use the contrast between personas, apply a strong strategy, and present their reasoning clearly and logically.
The most promising plan is selected among the three, resulting in one best plan per persona. 
For each speaker persona, one finalized plan is thus determined and passed to the subsequent generation stage.
The prompts used for plan generation and selection can be found in Figure~\ref{fig:step_wise_generation_prompt} and Figure~\ref{fig:step_wise_selection_prompt}, respectively.

\paragraph{Step 4: Counterargument Generation and Selection}
Building on the selected plans, the finalized plan for each speaker persona is explicitly used, along with the original post, OP persona, and the designated speaker persona, to guide counterargument generation.
The LLM then generates three candidate counterarguments per persona, each following the selected plan while reflecting the persona’s distinct perspective.
A voting procedure evaluates these candidates based on whether they leverage the contrast between personas, directly challenge the original argument, and are specific, persuasive, and logically consistent.
Among the three candidates, the most persuasive counterargument is selected as the final output for that persona. 
This process is repeated for each persona, with one best counterargument selected per persona, yielding three diverse counterarguments from the same, nearest, and furthest clusters that collectively capture multiple perspectives.
The prompts for this step are identical to those in Step~3, shown in Figure~\ref{fig:step_wise_generation_prompt} and Figure~\ref{fig:step_wise_selection_prompt}, respectively.

\section{Experiments}
\label{sec:experiments}
\subsection{Dataset}
\label{sec:dataset}

To assess the ability to generate diverse and persuasive counterarguments, we obtained the dataset of 847 \emph{ChangeMyView} (CMV) subreddit\footnote{\url{https://www.reddit.com/r/changemyview/}} posts (i.e., arguments) from Academic Torrents\footnote{\url{https://academictorrents.com/details/1614740ac8c94505e4ecb9d88be8bed7b6afddd4}}, each paired with three comments (i.e., counterarguments) that have successfully persuaded the original poster as gold-standard persuasive counterarguments.
On CMV, each post consists of a title summarizing the main claim and a body providing supporting premises.
Original posters award a delta ($\Delta$) to comments that successfully change their view, which we treat as quality counterarguments.
We first collected 72,999 posts from CMV, spanning the years 2013 to 2023.
Then, to ensure diversity, we filter for posts with three delta-awarded comments, yielding a dataset of 847 post-comments pairs.




\subsection{Baselines}
\label{sec:baseline_models}
To establish meaningful comparisons, we adopt baselines from both prior counterargument generation research and representative LLMs.

\paragraph{Backbone Language Models}

\begin{itemize}
    \item \textbf{LLaMA3.1-8B}~\citep{grattafiori2024llama}. 
An instruction-tuned model from the LLaMA3.1 family trained on large-scale publicly available datasets. 
In particular, we use \texttt{LLaMA3.1-8B-Instruct} as the primary backbone for our proposed method.
    \item \textbf{LLaMA3.2-1B}~\citep{grattafiori2024llama}. 
A more lightweight variant of the LLaMA family designed for efficient inference. 
We employ the \texttt{LLaMA3.2-1B-Instruct} model to evaluate performance on a significantly smaller parameter scale.
    \item \textbf{Qwen3-8B} and \textbf{Qwen-14B}~\citep{qwen3technicalreport}.
A family of state-of-the-art large language models with strong reasoning and language capabilities. 
We evaluate both \texttt{Qwen3-8B} and \texttt{Qwen3-14B} models in non-thinking mode to provide comparisons across different architectures and model sizes.
\end{itemize}




\paragraph{Counterargument Generation Methods}

\begin{itemize}
    \item \textbf{Vanilla Generation (Vanilla)}.
In this setting, the model is prompted to directly produce three counterarguments in a single pass, as illustrated in Figure~\ref{fig:base_llm_prompt}.
    \item \textbf{Argument Undermining}~\citep{alshomary-etal-2021-counter}. 
This method identifies weak premises in the original post and generates counterarguments by attacking them. 
For fair comparison, we first employ a weak-premise identification model to select the top three weak premises. 
We then generate one counterargument per premise, resulting in three outputs.
    \item \textbf{Joint One-seq}~\citep{alshomary-wachsmuth-2023-conclusion}. 
This method infers multiple conclusions from the premises in the original post and uses them as the basis for counterargument generation. 
Following the original formulation, we sample three conclusions and generate a counterargument for each.
\end{itemize}

\subsection{Evaluation Metrics}
\label{sec:evaluation_metrics}
\begin{table*}[t!]
\centering
{%
\small
\setlength{\tabcolsep}{5.5pt}
\begin{tabular}{l l r r r}
\toprule
\multirow{2}{*}{\textbf{Backbone}} 
& \multirow{2}{*}{\textbf{Method}}
& \multicolumn{2}{c}{\textbf{\textit{Persuasiveness}}}  
& \multicolumn{1}{c}{\textbf{\textit{Perspective}}} \\
\hhline{~~ -- ~}
& 
&
\multicolumn{1}{c}{\textbf{General}} &
\multicolumn{1}{c}{\textbf{Targeted}} &
\multicolumn{1}{c}{\textbf{\textit{Diversity}}} \\
\midrule
\multirow{4}{*}{\texttt{LLaMA3.1-8B}}
& Vanilla 
& $7.79_{\pm 0.0031}$ & $7.59_{\pm 0.0023}$ 
& $3.78_{\pm 0.0021}$ \\
& CoT
& $7.18_{\pm 0.0036}$ & $6.96_{\pm 0.0043}$ & $3.51_{\pm 0.0095}$ \\
& Argument Undermining 
& $4.23_{\pm 0.0040}$ & $4.34_{\pm 0.0021}$
& $2.76_{\pm 0.0015}$ \\
& Joint One-seq 
& $4.62_{\pm 0.0055}$ & $4.69_{\pm 0.0067}$
& $3.15_{\pm 0.0066}$ \\
& \method (Ours)
& $\mathbf{8.32}_{\pm 0.0025}$ & $\mathbf{7.93}_{\pm 0.0032}$
& $\mathbf{4.28}_{\pm 0.0024}$ \\
\midrule

\multirow{4}{*}{\texttt{LLaMA3.2-1B}}
& Vanilla 
& $6.63_{\pm 0.0031}$ & $6.44_{\pm 0.0039}$
& $3.34_{\pm 0.0032}$ \\
& CoT
& $5.31_{\pm 0.0063}$ & $4.95_{\pm 0.0046}$ & $2.56_{\pm 0.0023}$ \\
& Argument Undermining 
& $3.95_{\pm 0.0034}$ & $3.92_{\pm 0.0040}$
& $2.49_{\pm 0.0043}$ \\
& Joint One-seq 
& $3.56_{\pm 0.0056}$ & $3.44_{\pm 0.0027}$
& $2.54_{\pm 0.0048}$ \\
& \method (Ours)
& $\mathbf{7.11}_{\pm 0.0022}$ & $\mathbf{6.84}_{\pm 0.0016}$
& $\mathbf{3.61}_{\pm 0.0029}$ \\
\midrule

\multirow{4}{*}{\texttt{Qwen3-8B}}
& Vanilla 
& $8.18_{\pm 0.0024}$ & $7.94_{\pm 0.0029}$
& $3.59_{\pm 0.0069}$ \\
& CoT
& $7.93_{\pm 0.0035}$ & $7.73_{\pm 0.0018}$ & $4.03_{\pm 0.0065}$ \\
& Argument Undermining 
& $5.54_{\pm 0.0033}$ & $5.59_{\pm 0.0041}$
& $2.13_{\pm 0.0051}$ \\
& Joint One-seq 
& $3.81_{\pm 0.0009}$ & $3.80_{\pm 0.0062}$
& $2.61_{\pm 0.0062}$ \\
& \method (Ours)
& $\mathbf{8.72}_{\pm 0.0026}$ & $\mathbf{8.22}_{\pm 0.0025}$
& $\mathbf{4.71}_{\pm 0.0070}$ \\
\midrule

\multirow{4}{*}{\texttt{Qwen3-14B}}
& Vanilla 
& $8.33_{\pm 0.0022}$ & $8.02_{\pm 0.0031}$
& $4.19_{\pm 0.0051}$ \\
& CoT
& $8.24_{\pm 0.0013}$ & $7.97_{\pm 0.0021}$ & $4.16_{\pm 0.0069}$ \\
& Argument Undermining 
& $5.38_{\pm 0.0016}$ & $5.56_{\pm 0.0030}$
& $2.17_{\pm 0.0035}$ \\
& Joint One-seq 
& $4.10_{\pm 0.0060}$ & $4.16_{\pm 0.0027}$
& $2.79_{\pm 0.0024}$ \\
& \method (Ours)
& $\mathbf{8.73}_{\pm 0.0031}$ & $\mathbf{8.29}_{\pm 0.0008}$
& $\mathbf{4.28}_{\pm 0.0062}$ \\

\bottomrule
\end{tabular}
}
\caption{Evaluation results of baselines and \method.
All scores are averaged over five runs and reported with standard deviations.
\textit{Persuasiveness} is reported for general and targeted settings (1--10 scale), 
\textit{Perspective diversity} evaluates the variety of viewpoints reflected in generated counterarguments (1--5 scale).
}
\label{tab:llm_evaluation_main_table}
\end{table*}

We evaluate our method across two dimensions: Persuasiveness and Perspective Diversity.
We employ \emph{LLM-as-a-Judge}~\citep{zheng2023judging, gu2024survey} using GPT-4o-mini, as traditional lexical metrics like BLEU and ROUGE often diverge from human judgment~\citep{celikyilmaz2020evaluation, hu-etal-2024-americano-argument}.
For targeted persuasiveness, which measures how well the counterarguments resonate with the original poster(OP), we additionally use data-driven \emph{classifier-based scores}.

\noindent\textbf{Persuasiveness}
While general persuasiveness is important, it is equally crucial for counterarguments to be persuasive with respect to the OP’s context.
To capture both, we evaluate \emph{general persuasiveness} using only the title (claim) of the original post, and \emph{targeted persuasiveness} using both the title and body (premises), which assesses resonance with the original poster’s context and simulates realistic dialogue.
Since each input yields multiple counterarguments, we report the average persuasiveness score to reflect overall persuasiveness.
The prompts used for these two evaluations are provided in Appendix Figures~\ref{fig:general_persuasiveness_prompt} and \ref{fig:targeted_persuasiveness_prompt}.

\noindent\textbf{Perspective Diversity}
Producing counterarguments that embody genuinely distinct viewpoints, beyond superficial rewordings, is difficult.
True diversity requires capturing ideological, emotional, or experiential variation that reflects distinct perspectives.
We therefore evaluate whether the outputs move beyond lexical variation and invoke deeper interpretive frames not explicitly in the input.
The prompt for evaluating perspective diversity is provided in Appendix Figure~\ref{fig:perspective_diversity_prompt}.

\section{Results and Analysis}
\label{sec:main_results}

\subsection{Rating-scale Evaluation}
\label{sec:llm-based_evaluation}

The reported scores were averaged across five runs to mitigate random variation.
As shown in Table~\ref{tab:llm_evaluation_main_table}, \method consistently outperforms baseline methods across key dimensions, persuasiveness and perspective diversity, when compared within the same backbone model.
The persuasiveness scores indicate that \method-generated counterarguments are more effective for both a general audience and OP.

While standard Chain-of-Thought (CoT) prompting serves as a strong general baseline, it falls short in targeted persuasiveness, indicating that linear step-by-step reasoning alone struggles to capture individual recipient contexts.
Other baselines, such as Argument Undermining and Joint One-seq, also show a significant drop in persuasiveness.
In contrast, \method maintains strong performance, surpassing the Vanilla baseline while clearly improving perspective diversity.
These results show that \method is a robust and scalable solution that effectively improves both the quality and the variety of perspectives across different model scales.


\subsection{Classifier-based Evaluation}
\label{sec:classifier-based_evaluation}

\begin{table}[t!]
\small
\centering
\resizebox{\columnwidth}{!}{
    \begin{tabular}{l l r r}
    \toprule
    \textbf{Backbone} & \textbf{Method} & \textbf{Avg.} & \textbf{Max}\\
    \midrule
    
    \multirow{4}{*}{\texttt{LLaMA3.1-8B}}
    & Vanilla        & \textbf{0.84}  & 0.88\\
    & CoT           & 0.82 & 0.87 \\
    & Argument Undermining        & 0.64 & 0.83 \\
    & Joint One-seq        & 0.42 & 0.78 \\
    & \method (Ours) & 0.81 & \textbf{0.90} \\
    \midrule
    
    \multirow{4}{*}{\texttt{LLaMA3.2-1B}}
    & Vanilla        & 0.80 & 0.86 \\
    & CoT              & 0.79 & 0.86 \\
    & Argument Undermining        & 0.69 & 0.87 \\
    & Joint One-seq        & 0.32 & 0.74 \\
    & \method (Ours) & \textbf{0.82} & \textbf{0.88} \\
    \midrule

    \multirow{4}{*}{\texttt{Qwen3-8B}}
    & Vanilla        & 0.83 & 0.86 \\
    & CoT           & 0.82 & 0.85 \\
    & Argument Undermining        & 0.84 & 0.90 \\
    & Joint One-seq        & 0.49 & 0.79 \\
    & \method (Ours) & \textbf{0.88} & \textbf{0.91} \\
    \midrule

    \multirow{4}{*}{\texttt{Qwen3-14B}}
    & Vanilla        & 0.85 & 0.88 \\
    & CoT           & 0.85 & 0.88 \\
    & Argument Undermining        & 0.81 & 0.88 \\
    & Joint One-seq        & 0.55 & 0.81 \\
    & \method (Ours) & \textbf{0.88} & \textbf{0.90} \\
    \midrule

    \end{tabular}
    
}
\caption{Targeted persuasiveness is evaluated using the delta classifier. 
The score ranges from –1 to 1 and represents the predicted probability of receiving a delta.
}
\label{tab:delta_score_table}
\end{table}

To complement LLM judgments, we evaluate targeted persuasiveness using a trained Delta classifier (evaluator performance in Table~\ref{tab:delta_classifier_results}).
As presented in Table~\ref{tab:delta_score_table}, \method demonstrates strong performance across all backbones, consistently outperforming prior counterargument generation methods.

Notably, \method achieves the best Avg and Max scores on most backbones, including LLaMA3.2-1B, Qwen3-8B, and Qwen3-14B, indicating its ability to generate highly persuasive counterarguments.
For LLaMA3.1-8B, although the Avg score is slightly lower than the Vanilla baseline, \method attains the highest Max score (0.90), suggesting that it can produce particularly persuasive arguments among its generated candidates.
These results confirm that \method effectively improves targeted persuasiveness across diverse backbone models.

\begin{figure*}[t!]
\centering
\includegraphics[width=\textwidth]{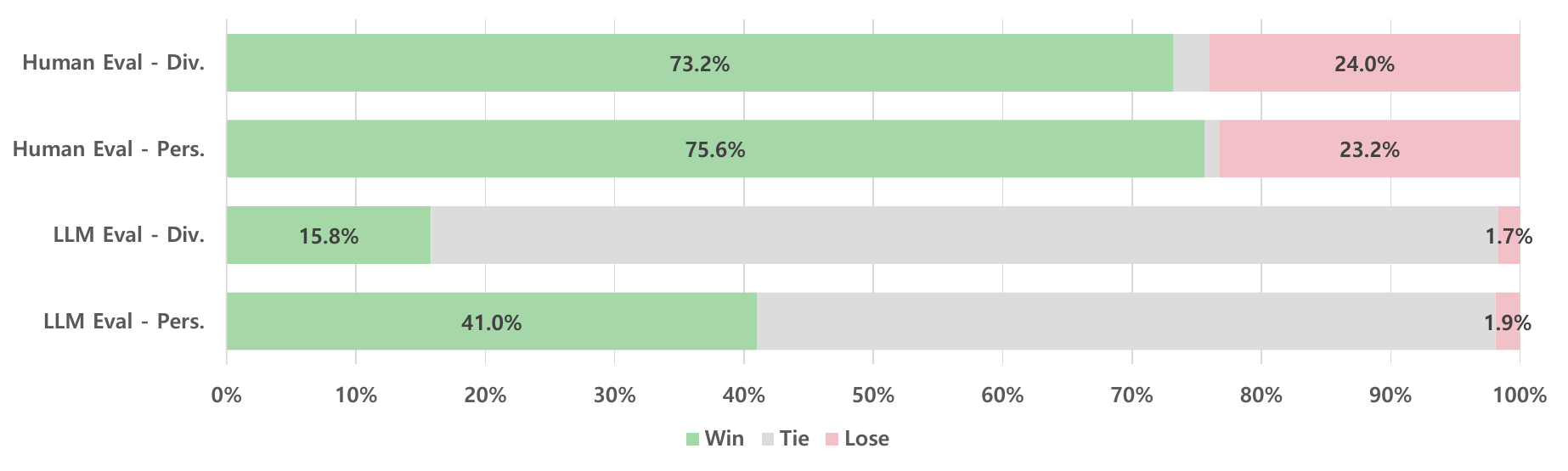}
\caption{Win/Tie/Lose analysis of diversity (Div.) and persuasiveness (Pers.) across Human Evaluation (top) and LLM-based Evaluation (bottom).
The figure compares our method (\method) with baselines using LLaMA3.1-8B as the main backbone model, showing the proportion of cases where each system’s counterarguments were judged as more diverse or persuasive (Win), equivalent (Tie), or less effective (Lose).}
\label{fig:win_tie_lose_analysis}
\end{figure*}


\subsection{Pairwise Evaluation}
\label{sec:llm-based_pairwise_evaluation}


To evaluate our framework, we conduct human and LLM-based pairwise comparisons on counterargument persuasiveness and diversity.

For the human evaluation, we randomly sampled 95 inputs\footnote{This evaluation scale is larger than those in prior studies~\citep{hu-etal-2025-debate, alshomary-wachsmuth-2023-conclusion}.} and asked five human evaluators to rate the outputs from \method and the baselines.
Details are in Appendix Section~\ref{sec:detailed_human_evaluation}.
As shown in Figure~\ref{fig:win_tie_lose_analysis} (top), evaluators consistently rated \method higher than the baseline in both diversity and persuasiveness.\footnote{Inter-annotator agreement, assessed via Fleiss’ Kappa (0.619) and Krippendorff’s Alpha (0.612), indicates substantial consistency~\citep{Landis77} among annotators.} Since both systems rely on the same LLaMA3.1-8B backbone model, this comparison isolates the effect of the framework itself.
In terms of diversity, the high win rate demonstrates that \method effectively avoids repetitive reasoning by exploring multifaceted logical paths.
Regarding persuasiveness, evaluators found \method significantly more convincing, indicating that the framework generates structurally sound, highly persuasive arguments.
Evaluators noted that \method provides calm, evidence-based explanations with concrete examples instead of a confrontational tone.
Although baseline models were noted for their directness, \method was consistently favored for integrating clear claims with compelling reasoning.

We also conducted pairwise comparisons using GPT-4o-mini~\citep{OpenAI2024-4o-mini} as the evaluator.
To mitigate potential ordering effects~\citep{zheng2023judging}, we presented the counterargument sets in both orders\footnote{Each evaluation pair was tested in both AB and BA sequences, and the final results were aggregated accordingly.}.
We also allowed for \textit{Tie}, which captures cases where the evaluator either explicitly selected a ``Hard'' option or produced inconsistent preferences when the order of the two sets was reversed, reflecting potential ordering effects inherent in LLM-based pairwise judgments.
Figure~\ref{fig:win_tie_lose_analysis} (bottom) presents the pairwise comparison results between \method and its Vanilla version using LLaMA3.1-8B as the backbone model.
LLM-based pairwise evaluations tend to produce relatively high tie rates due to potential ordering bias.
Nevertheless, the results show that \method achieves a higher proportion of wins in both diversity and persuasiveness, further supporting the improvements observed in the human evaluation.
Additional LLM-based evaluation results across other backbone models are provided in Appendix Section~\ref{sec:additional_llm_eval}.
These results demonstrate that \method consistently improves both perspective diversity and persuasiveness compared to the vanilla baseline across all evaluations.

\subsection{Qualitative Analysis}
\label{sec:qualitative_analysis}
\begin{table*}[!t]
\small
\centering
{
\setlength{\tabcolsep}{5.5pt}
\begin{tabular}{l r r r}
\toprule
\multirow{2}{*}{\textbf{Configuration}} 
& \multicolumn{2}{c}{\textbf{\textit{Persuasiveness}}}  
& \multicolumn{1}{c}{\textbf{\textit{Perspective}}}   \\
\hhline{~ --} 
& \multicolumn{1}{c}{\textbf{General}}  
& \multicolumn{1}{c}{\textbf{Targeted}}  
& \multicolumn{1}{c}{\textbf{\textit{Diversity}}}  \\
\midrule
\texttt{LLaMA3.1-8B} &
$7.79_{\pm 0.0031}$ & $7.59_{\pm 0.0023}$ 
& $3.78_{\pm 0.0021}$ \\
\quad + Persona &
$7.79_{\pm 0.0014}$ & $7.16_{\pm 0.0050}$ &
$4.24_{\pm 0.0040}$ \\
\quad + CoT-based Gen. ($n$=1; linear chain) &
$7.98_{\pm 0.0015}$ & $7.75_{\pm 0.0018}$ &
$1.99_{\pm 0.0097}$ \\
\quad + Tree-based Gen. ($n$=3; tree search) &
$8.01_{\pm 0.0019}$ & $7.77_{\pm 0.0024}$ &
$2.25_{\pm 0.0075}$ \\
\quad + Persona + Tree-based Gen. (\method) &
$\mathbf{8.32}_{\pm 0.0025}$ & $\mathbf{7.93}_{\pm 0.0032}$
& $\mathbf{4.28}_{\pm 0.0024}$ \\
\bottomrule
\end{tabular}
}
\caption{Ablation study of \method. 
All scores are averaged over five runs and reported with standard deviations.
}
\label{tab:ablation_all}
\end{table*}

Analysis of generated counterarguments\footnote{Please refer to Table~\ref{tab:comparison_ex}, Table~\ref{tab:comparison_ex_2}, and Table~\ref{tab:persona_case_study} for examples.} shows that \method moves beyond abstract ethical reasoning by adding richer social and practical contexts.

In the organ donation debate, baseline outputs largely repeat general arguments about autonomy and utilitarianism.
In contrast, \method grounds its counterarguments in concrete considerations, such as societal trust in healthcare and state control over bodily decisions.
This framing makes the arguments more relatable and easier to evaluate in real-world terms.
Similarly, in the scientific theory patenting example, the baseline focuses on a single claim about innovation, whereas \method introduces multifaceted perspectives through diverse personas to produce more nuanced and persuasive responses.

Furthermore, qualitative analysis reveals distinct reasoning patterns across persona types classified by embedding-based semantic distance.
\textit{Same} personas target internal contradictions within the original focus, \textit{Nearest} personas introduce relevant contextual nuances, and \textit{Furthest} personas fundamentally reframe the debate.
To illustrate this spectrum in practice, Table~\ref{tab:persona_case_study} presents a detailed case study on restricting children's late-night movie access.
Here, a film scholar reinterprets the cinematic experience, a literature scholar critiques the demand for silence as elitist, and a real estate developer shifts the focus entirely to local business impacts.
This shows that using personas at different distances helps the model provide both sharp logical critiques and broader, new perspectives.

Overall, \method achieves superior quality by grounding arguments in practical, context-rich reasoning.
Through its multi-step reasoning process, our framework systematically enhances both the depth and structural completeness of generated arguments.
These examples clearly demonstrate how persona conditioning shapes the reasoning style of each counterargument.

\subsection{Ablation Study}
\label{sec:ablation_study}
We analyze the contribution of each component through ablation experiments, which evaluate each module individually and in combination.
Starting from the baseline LLaMA3.1-8B model, we incrementally introduce persona grounding, CoT-based, and tree-based generation.

Persona grounding alone substantially boosts perspective diversity by introducing various social backgrounds and viewpoints.
However, it fails to improve persuasiveness and even leads to a decrease in targeted persuasiveness, suggesting that diverse viewpoints alone do not guarantee a compelling argument.
In contrast, compared to the LLaMA 3.1 baseline, tree-based generation contributes most to persuasiveness by exploring multiple reasoning paths.
However, this structured search process leads to a sharp decline in perspective diversity.
This indicates that while the search process optimizes for argumentative strength, it tends to converge on a narrow set of logical paths.
Ultimately, \method, which integrates both persona grounding and tree-based generation, achieves the highest scores across all metrics.
It successfully recovers the diversity lost during the tree search, while simultaneously attaining the highest levels of both general and targeted persuasiveness.
This demonstrates that \method effectively translates diverse persona-driven signals into robust persuasive arguments through structured reasoning.

\section{Conclusion}
\label{sec:conclusion}
In this work, we proposed and addressed the task of generating multiple distinct and persuasive counterarguments grounded in realistic personas.
Inspired by the Tree-of-Thoughts approach, we proposed a tree-based stepwise generation with a pruning process to improve reasoning quality, while integrating personas grounding through distance-based persona selection.
This design enables our framework to explore diverse argumentative perspectives while maintaining a strong persuasive structure, overcoming the inherent limitations of standard LLM generation.
For evaluation, we conducted comprehensive experiments combining human evaluation, LLM-based judgments, and classifier-based assessments.
The results demonstrate that our persona-grounded and tree-based generation approach significantly improves both the diversity and persuasiveness of counterarguments.
Our work highlights the importance of structured reasoning and persona-driven perspective-taking for counterargument generation, suggesting a promising direction for building LLM systems that support critical thinking, balanced debates, and informed decision-making.

\section*{Ethical Considerations}
\label{sec:ethics_statement}
This study makes use of the publicly available Reddit ChangeMyView dataset, which has also been widely adopted in prior work~\citep{alshomary-wachsmuth-2023-conclusion, lin-etal-2023-argue} on computational argumentation.
The dataset was used strictly for research purposes.
We are aware that generating persuasive counterarguments with large language models raises ethical concerns, particularly the risk of misuse in manipulative or coercive contexts.
To reduce these risks, our work is limited to academic exploration, with the goal of examining diversity of perspectives rather than promoting adversarial persuasion.

In addition, to prevent the risk of outputs being mistaken for human speech in deceptive ways, we restricted the use of first-person generation.
To protect the privacy of the original authors, we anonymized all user-identifiable information by replacing original usernames with generic identifiers (e.g., User 1, User 2).
Furthermore, we paraphrased the examples cited in this paper while preserving their semantic intent. 
These modifications were applied to the illustrative examples, whereas the actual computational analysis was conducted on the original anonymized data.
Thus, our study focused on the substantive content and patterns of the arguments rather than the identities of individual authors.

Additionally, our reliance on English-based Reddit data may introduce demographic biases and limit the generalizability of our findings to other languages or cultural rhetorical traditions.
We also emphasize the need for future work to consider safeguards and responsible use practices when deploying such systems.

\section*{Limitations}
\label{sec:limitations}
While \method demonstrates strong performance, we identify several limitations that offer promising avenues for future work.
First, this work focuses on generating counterarguments for a single opinion.
However, debates often unfold over multiple rounds, with each round consisting of an exchange between participants~\citep{durmus-cardie-2019-corpus,li-etal-2020-exploring-role}.
Extending this framework to multi-round debates would enable models to engage in interactive and dynamic exchanges, where counterarguments are refined over multiple turns. 
This line of research could further extend to multi-party debates, where multiple participants interact and compete~\citep{sia-etal-2022-offer}.
Second, the current framework is limited to text-based personas derived from clustering and relies on automated LLM planning without real-time human intervention.
While this setup maximizes operational efficiency, it can introduce potential LLM biases.
To address this, incorporating richer user signals (such as value-based attributes~\citep{lukin-etal-2017-argument}) alongside human feedback mechanisms like RLHF could improve the persuasiveness of generated counterarguments, while also offering finer-grained control over model biases.


\section*{Acknowledgments}
We would like to thank the anonymous reviewers for their helpful questions and comments.
This work was partly supported by Institute of Information \& communications Technology Planning \& Evaluation(IITP) grant funded by the Korea government(MSIT)
(RS-2019-II190421, Artificial Intelligence Graduate School Program (Sungkyunkwan University) \& 
RS-2025-02263169, Detection and Prediction of Emerging and Undiscovered Voice Phishing \&
RS-2025-25442569, AI Star Fellowship Support Program(Sungkyunkwan University))
 and the Ministry of Education of the Republic of Korea and the National Research Foundation of Korea (NRF-RS-2025-00523385 \& RS-2024-00333484).


\bibliography{custom,anthology-1}


\clearpage
\appendix

\section{Appendix}
\label{sec:appendix}
\renewcommand\thesubsection{A\arabic{subsection}}

\setcounter{table}{0}
\renewcommand{\thetable}{A\arabic{table}}

\setcounter{figure}{0}
\renewcommand{\thefigure}{A\arabic{figure}}

\subsection{LLM Usage}
We employed GPT-4o to support the manuscript preparation process. 
Specifically, the model was used to assist in identifying relevant prior work and refining the clarity and readability of the draft. 
This usage was limited to auxiliary scholarly support—such as improving grammar and reducing stylistic inconsistencies.

\subsection{Persona Cluster Examples}
\label{sec:cluster-examples}

To illustrate how the persona clusters are organized, we present representative examples from three clusters: (1) Sports and Physical Education, (2) Finance and Marketing, and (3) Historians.
These examples demonstrate the semantic coherence within each cluster, reflecting domain-specific interests and professional backgrounds.

\paragraph{Sports and Physical Education}
\begin{itemize}[leftmargin=*, nosep]
    \item A high school physical education teacher seeking to incorporate Paralympic history and achievements into the curriculum to inspire and educate students about inclusivity in sports.
    \item A sports scientist researching the biomechanics and physics of tennis, focusing on how racket specifications impact performance and injury risks.
    \item A sports journalist covering the history of ice hockey and its impact on national identity in Poland.
    \item An elementary school teacher who enjoys incorporating diverse sports stories in her curriculum to inspire students.
    \item A football coach seeking to learn from successful strategies and team management in various leagues.
\end{itemize}

\paragraph{Finance and Marketing}
\begin{itemize}[leftmargin=*, nosep]
    \item A financial analyst specializing in Asian markets and wealthy individuals, interested in tracking the investments and philanthropic activities of billionaires like Gerald Chan.
    \item A quantitative analyst with expertise in financial modeling and algorithmic trading, seeking to develop and implement systematic value investment strategies.
    \item A digital marketing specialist interested in innovative aggregator models that consolidate search results from multiple sources.
    \item A marketing specialist for a tech company, looking for innovative ways to engage with pop culture and fandoms to promote new products and services.
    \item A business strategist for Arriva UK Bus, interested in exploring opportunities and challenges related to subsidiary operations and company restructuring.
\end{itemize}

\paragraph{Historians}
\begin{itemize}[leftmargin=*, nosep]
    \item An Iowa historian focusing on the development and growth of townships in Jones County.
    \item A historian specializing in 19th-century British architecture, with a focus on the works of notable architects in Lancashire.
    \item A local historian specializing in the political and business development of Marlborough, Massachusetts in the 19th century.
    \item A historian specializing in the late medieval and early modern history of France and the Iberian Peninsula, with a focus on power dynamics, family strategies, and women's roles in politics.
    \item A local historian or genealogist researching the history of small communities and families in Fremont County, Iowa.
\end{itemize}

\subsection{Qualitative Analysis of Estimated Personas}

To evaluate the quality of the estimated personas for subsequent steps in \method, we conducted a qualitative analysis of 100 randomly sampled post and estimated persona pairs. We focused on whether the generated personas were consistent with the posts, whether they captured the author's characteristics, and whether they introduced overgeneralization. The estimated personas consistently captured both explicit self-disclosed information and the authors' perspectives. This suggests that the estimated personas provide useful information for both plan generation and counterargument generation.

In most cases, the model effectively extracted author’s personal attributes, core interests and viewpoints. When posts contained specific personal attributes, such as age, sex, or organizational affiliation, the model accurately captured these details. Beyond simply extracting explicit attributes, the model effectively synthesized broader contextual cues, such as recurring interests, personal experiences, and expressed viewpoints. For instance, posts reflecting personal deliberations were summarized into personas that reflected the author's underlying values and decision-making criteria, while discussions on specific experiences or hobbies were captured as clear indicators of topic familiarity and personal interest.

Although a few generated personas included implicit details not explicitly stated in the text, this does not undermine their utility. Our goal is not to strictly reconstruct the author's real-world identity, but to infer plausible persona signals that unlock different perspectives for counterargument generation. As demonstrated in Table~\ref{tab:persona_case_study}, whether the inferred persona closely aligns with the original setting (e.g., a film scholar) or reframes it from a distinct domain (e.g., a real estate developer), these signals reliably guide the model to produce contextually rich and highly relevant counterarguments.

\subsection{Implementation Details}
Our models were developed and evaluated using multiple GPU resources.
Specifically, we used an NVIDIA A100 GPU (80GB) to implement the argument undermining and joint one-seq methods.
For inference and testing, the vanilla baseline and our \method were deployed on two NVIDIA RTX A6000 GPUs (48GB each) using vLLM (v0.8.5) \citep{kwon2023efficient} to ensure efficient processing. 
During the generation process, we set the temperature to 0.8 and top-p to 0.95 to balance diversity and coherence, with the maximum token length limited to 2,048.
Additionally, API-based models, including gpt-4o-mini and text-embedding-3-large, were accessed via OpenAI’s API.

\subsection{LLM-based Evaluation: Quality}
\label{sec:llm-based_evaluation_quality}

\begin{table}[t!]
\centering
\resizebox{\columnwidth}{!}{%
\small
\setlength{\tabcolsep}{5.5pt}
\begin{tabular}{l l rrrr}
\toprule
\multicolumn{1}{c}{\textbf{Backbone}} &
\multicolumn{1}{c}{\textbf{Method}} &
\multicolumn{1}{c}{\textbf{App.}} &
\multicolumn{1}{c}{\textbf{Cla.}} &
\multicolumn{1}{c}{\textbf{Gra.}} &
\multicolumn{1}{c}{\textbf{Rel.}}
\\
\midrule
\multirow{4}{*}{\texttt{LLaMA3.1-8B}}
& Vanilla 
& 4.40 & 4.27 & 4.74 & 4.70 \\
& AU
& 2.46 & 2.36 & 2.37 & 2.80 \\
& Joint One-seq 
& 3.02 & 3.06 & 3.22 & 3.22 \\
& \method (Ours)
& \textbf{4.55} & \textbf{4.34} & \textbf{4.87} & \textbf{4.75} \\
\midrule

\multirow{4}{*}{\texttt{LLaMA3.2-1B}}
& Vanilla 
& \textbf{3.95} & \textbf{3.90} & \textbf{4.32} & 4.18 \\
& AU 
& 2.55 & 2.46 & 2.61 & 2.59 \\
& Joint One-seq 
& 2.33 & 2.31 & 2.48 & 2.30 \\
& \method (Ours)
& 3.90 & 3.83 & 4.29 & \textbf{4.19} \\
\midrule

\multirow{4}{*}{\texttt{Qwen3-8B}}
& Vanilla 
& \textbf{4.71} & \textbf{4.55} & \textbf{4.94} & \textbf{4.90} \\
& AU
& 3.40 & 3.30 & 3.45 & 3.53 \\
& Joint One-seq 
& 2.42 & 2.40 & 2.50 & 2.59 \\
& \method (Ours)
& 4.28 & 4.27 & 4.84 & 4.63 \\
\midrule

\multirow{4}{*}{\texttt{Qwen3-14B}}
& Vanilla 
& \textbf{4.79} & \textbf{4.62} & \textbf{4.96} & \textbf{4.94} \\
& AU
& 3.35 & 3.25 & 3.43 & 3.55 \\
& Joint One-seq 
& 2.61 & 2.60 & 2.68 & 2.81 \\
& \method (Ours)
& 4.70 & 4.60 & \textbf{4.96} &4.93 \\

\bottomrule
\end{tabular}
}
\caption{Evaluation results of baselines and \method measured by GPT-4o-mini. Reported scores have been averaged over five runs.
For \textit{quality}, we report appropriateness (App.), clarity (Cla.), grammaticality (Gra.), and relevance (Rel.), each on a 1--5 scale.}
\label{tab:llm_evaluation_quality_table}
\end{table}

Prior studies have established diverse criteria for evaluating the quality of arguments.
\citet{alshomary-etal-2021-counter} consider grammaticality and content richness as key factors in assessing generated arguments, while \citet{lin-etal-2023-argue} emphasize appropriateness, grammaticality, and logic, aligning GPT-based and human evaluation through shared criteria.
Moreover, \citet{wachsmuth-etal-2017-computational} provide a comprehensive taxonomy of argument quality, highlighting clarity, appropriateness, and Relevance as core components under effectiveness and reasonableness.
Based on these findings, we adopt four criteria for LLM-based evaluation of counterarguments: \\
\noindent\textbf{Appropriateness:} Whether the language and tone are suitable for the context and proportional to the issue. Inappropriate tone (e.g., overly aggressive or dismissive) lowers the score. \\
\noindent\textbf{Clarity:} Whether the writing is clear, well-organized, and free from ambiguity or unnecessary complexity, allowing the reader to easily follow the reasoning. \\
\noindent\textbf{Grammaticality:} Whether the text follows standard grammar conventions, including punctuation, sentence structure, and syntax. This ensures the counterargument reads fluently without errors. \\
\noindent\textbf{Relevance:} How directly the counterargument engages with the original post and addresses its key points. Superficial or off-topic content would reduce relevance. \\

\subsection{Quality Scores Across Backbone Models}
As presented in Table~\ref{tab:llm_evaluation_quality_table}, \method demonstrates robust performance in terms of generation quality across various backbone models.
Overall, \method consistently maintains high scores across all four dimensions.
While minor fluctuations in scores are observed compared to the Vanilla baseline, these variations remain within a highly acceptable range.
Such marginal differences are interpreted as a natural trade-off arising from the model's effort to adhere to specific persona constraints while maintaining linguistic universality.

\begin{table}[!ht]
\centering
\resizebox{\columnwidth}{!}{%
\setlength{\tabcolsep}{6pt}
\begin{tabular}{l r r r}
\toprule
\multirow{2}{*}{\textbf{Method / Configuration}} 
& \multicolumn{2}{c}{\textbf{\textit{Persuasiveness}}} 
& \multicolumn{1}{c}{\textbf{\textit{Perspective}}} \\
\hhline{~--~}
& \multicolumn{1}{c}{\textbf{General}} 
& \multicolumn{1}{c}{\textbf{Targeted}} 
& \multicolumn{1}{c}{\textbf{\textit{Diversity}}} \\
\midrule
\texttt{Vanilla} & 7.79 $\pm$ 0.0031 & 7.59 $\pm$ 0.0023 & 3.78 $\pm$ 0.0021 \\
\texttt{Persona\_All\_Same} & 8.32 $\pm$ 0.0038 & 7.93 $\pm$ 0.0042 & 4.25 $\pm$ 0.0064 \\
\texttt{Persona\_All\_Nearest} & \textbf{8.39 $\pm$ 0.0030} & \textbf{7.98 $\pm$ 0.0007} & 4.27 $\pm$ 0.0071 \\
\texttt{Persona\_All\_Furthest} & 8.25 $\pm$ 0.0034 & 7.86 $\pm$ 0.0022 & 4.17 $\pm$ 0.0068 \\
\midrule
\texttt{Method (Ours)} & 8.32 $\pm$ 0.0025 & 7.93 $\pm$ 0.0032 & \textbf{4.28 $\pm$ 0.0024} \\
\bottomrule
\end{tabular}
}
\caption{Quantitative evaluation across persona distance configurations on LLaMA3.1-8B. Scores are reported with standard deviations. Persuasiveness is evaluated on a 1--10 scale, and Perspective Diversity on a 1--5 scale.}
\label{tab:persona_distance_table}
\end{table}

\subsection{Impact of Persona Distance on Persuasiveness and Diversity}

To evaluate how persona distance affects counterargument generation, we compared three cluster configurations relative to the original post (OP) author: \texttt{Same}, \texttt{Nearest}, and \texttt{Furthest}. As shown in Table~\ref{tab:persona_distance_table}, persona distance creates a clear trade-off between targeted persuasiveness and perspective diversity, while overall argument quality remains stable.

These results align with key principles in social psychology. Counterarguments generated under \texttt{Same} and \texttt{Nearest} configurations achieved the highest targeted persuasiveness scores ($7.93$ and $7.98$, respectively). This supports the similarity-attraction effect~\citep{berscheid1966opinion}, which indicates that shared backgrounds and worldviews foster immediate trust and receptivity. By operating within the author's cognitive frame, closely aligned personas leverage arguments that directly resonate with the OP author.

Conversely, relying solely on a \texttt{Furthest} persona yields a lower targeted persuasiveness score ($7.86$). According to minority influence theory~\citep{nemeth1986differential}, exposure to contrasting viewpoints induces initial cognitive friction, making immediate agreement more challenging. Crucially, however, such distant perspectives stimulate divergent thinking. Rather than directly disputing the main claim, furthest personas reframe the issue by introducing alternative values, unconsidered edge cases, and new contextual angles. Ultimately, while proximal personas excel at direct persuasion, distant personas are essential for broadening perspective diversity, making their strategic integration a core strength of our framework.

\subsection{Ablation Study on Quality Scores}
The ablation study in Table~\ref{tab:ablation_quality_table} provides a deeper insight into how \method balances persona integration with output quality.
Our analysis reveals that simply injecting persona conditioning leads to a noticeable decline in quality scores, as the model prioritizes persona-specific traits over general clarity.
However, the integration of the tree-based step-wise generation and selection effectively mitigates this issue, successfully rebounding the scores back to levels comparable to the original LLaMA 3.1 baseline.
These results empirically demonstrate that the tree-based search mechanism serves as a critical quality-control layer, ensuring that the diversity gained through persona conditioning does not come at the expense of linguistic or logical integrity.
Consequently, \method achieves an optimal equilibrium between diverse perspectives and generation quality.

\begin{algorithm*}[t!]
\caption{Persona-guided Tree-based Counterargument Generation (\method)}
\label{algorithm:tree_of_personas}
\begin{algorithmic}[1]
\Require Original post $x$, persona cluster centroids $\mathcal{C}$ 
\Ensure Final counterarguments $\mathcal{Y} = \{y_1, y_2, y_3\}$

\State \textbf{Step 1: Original Poster (OP) Persona Estimation}
\State $p_{\text{op}} \gets \texttt{LLM\_Estimate\_Persona}(x)$
\Comment{Prompt: Figure \ref{fig:op_estimation_prompt}}
\State $c_{\text{op}} \gets \texttt{Nearest\_Centroid}(p_{\text{op}}, \mathcal{C})$

\State

\State \textbf{Step 2: Speaker Persona Selection}
\State Select three speaker personas from $\mathcal{C}$:
\Statex \hspace{1em} $p_{\text{same}} =$ centroid persona from $c_{\text{op}}$ \Comment{Same cluster}
\Statex \hspace{1em} $p_{\text{nearest}} =$ centroid persona from nearest cluster
\Statex \hspace{1em} $p_{\text{furthest}} =$ centroid persona from furthest cluster
\State $\mathcal{P}^\star = \{p_{\text{same}}, p_{\text{nearest}}, p_{\text{furthest}}\}$

\State

\State \textbf{Step 3: Plan Generation and Selection}
\For{each $p \in \mathcal{P}^\star$}
    \State $\{r_1, r_2, r_3\} \gets \texttt{LLM\_Generate\_Plans}(x, p_{\text{op}}, p)$
    \Comment{Prompt: Figure \ref{fig:step_wise_generation_prompt}}
    \State $r^\star_p \gets \texttt{LLM\_Select\_Plan}(\{r_1, r_2, r_3\}, p, p_{\text{op}})$
    \Comment{Prompt: Figure \ref{fig:step_wise_selection_prompt}}
\EndFor

\State

\State \textbf{Step 4: Counterargument Generation and Selection}
\For{each $p \in \mathcal{P}^\star$}
    \State $\{y^p_1, y^p_2, y^p_3\} \gets \texttt{LLM\_Generate\_Counters}(x, p_{\text{op}}, p, r^\star_p)$
    \Comment{Prompt: Figure \ref{fig:step_wise_generation_prompt}}
    \State $y^\star_p \gets \texttt{LLM\_Select\_Counter}(\{y^p_1, y^p_2, y^p_3\}, p, p_{\text{op}})$
    \Comment{Prompt: Figure \ref{fig:step_wise_selection_prompt}}
    \State Add $y^\star_p$ to $\mathcal{Y}$
\EndFor

\State

\State \Return $\mathcal{Y}$
\end{algorithmic}
\end{algorithm*}

\begin{table}[t!]
\small
\centering
\resizebox{\columnwidth}{!}{%
\setlength{\tabcolsep}{5.5pt}
\begin{tabular}{l rrrr}
\toprule
\multicolumn{1}{c}{\textbf{Configuration}} &
\multicolumn{1}{c}{\textbf{App.}} &
\multicolumn{1}{c}{\textbf{Cla.}} &
\multicolumn{1}{c}{\textbf{Gra.}} &
\multicolumn{1}{c}{\textbf{Rel.}} \\
\midrule
\texttt{LLaMA3.1-8B} &
4.40 & 4.27 & 4.74 & 4.70 \\
\quad + Persona &
3.82 & 3.81 & 4.38 & 3.99 \\
\quad + CoT-based Gen. ($n$=1; linear chain) &
4.69 & 4.43 & \textbf{4.93} & 4.82 \\
\quad + Tree-based Gen. ($n$=3; tree search) &
\textbf{4.70} & \textbf{4.44} & \textbf{4.93} & \textbf{4.83} \\
\quad + Persona + Tree-based Gen. (\method) &
4.55 & 4.34 & 4.87 & 4.75 \\
\bottomrule
\end{tabular}
}
\caption{Ablation study of \method.
For \textit{quality}, we report appropriateness (App.), clarity (Cla.), grammaticality (Gra.), and relevance (Rel.), each on a 1--5 scale.
}
\label{tab:ablation_quality_table}
\end{table}


\subsection{Human Evaluation}
\label{sec:detailed_human_evaluation}

\begin{figure}[t!]
\centering
\includegraphics[width=\columnwidth]{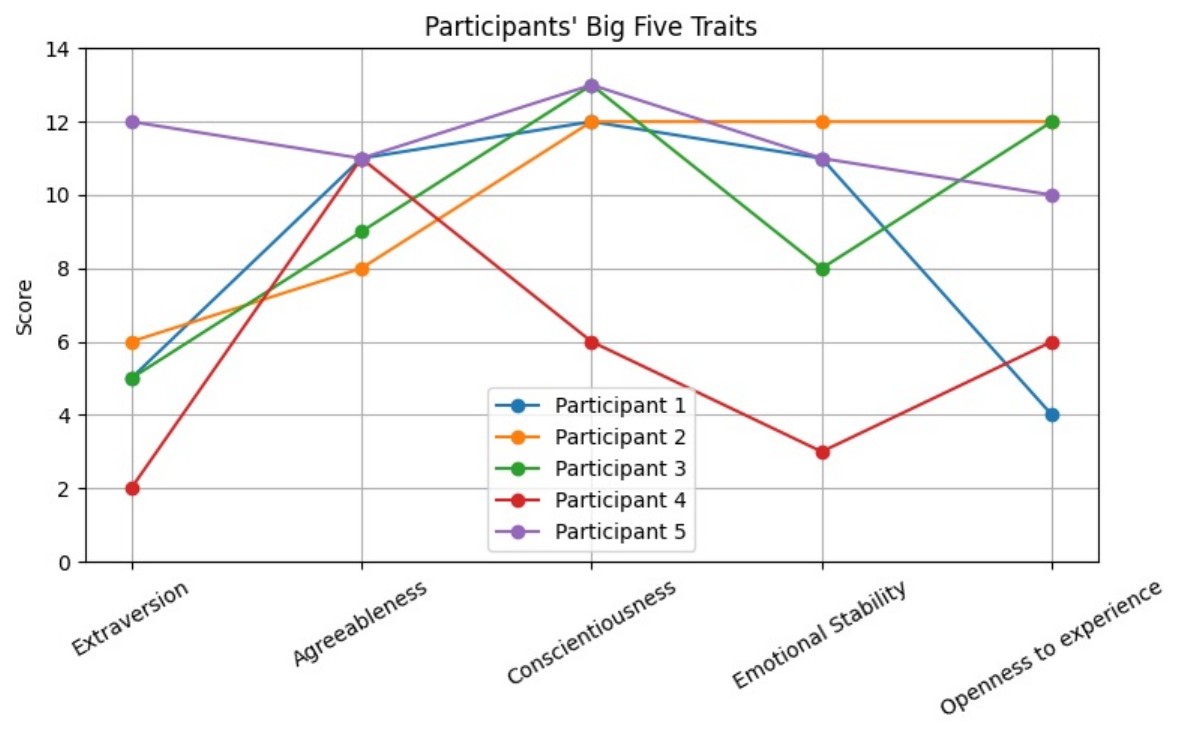}
\caption{Big Five Personality trait distribution of recruited evaluators. The heterogeneous profiles helped ensure diverse perspectives in human evaluation.}
\label{fig:human_eval_big_five}
\end{figure}

\begin{figure*}
\centering
\includegraphics[width=\textwidth]{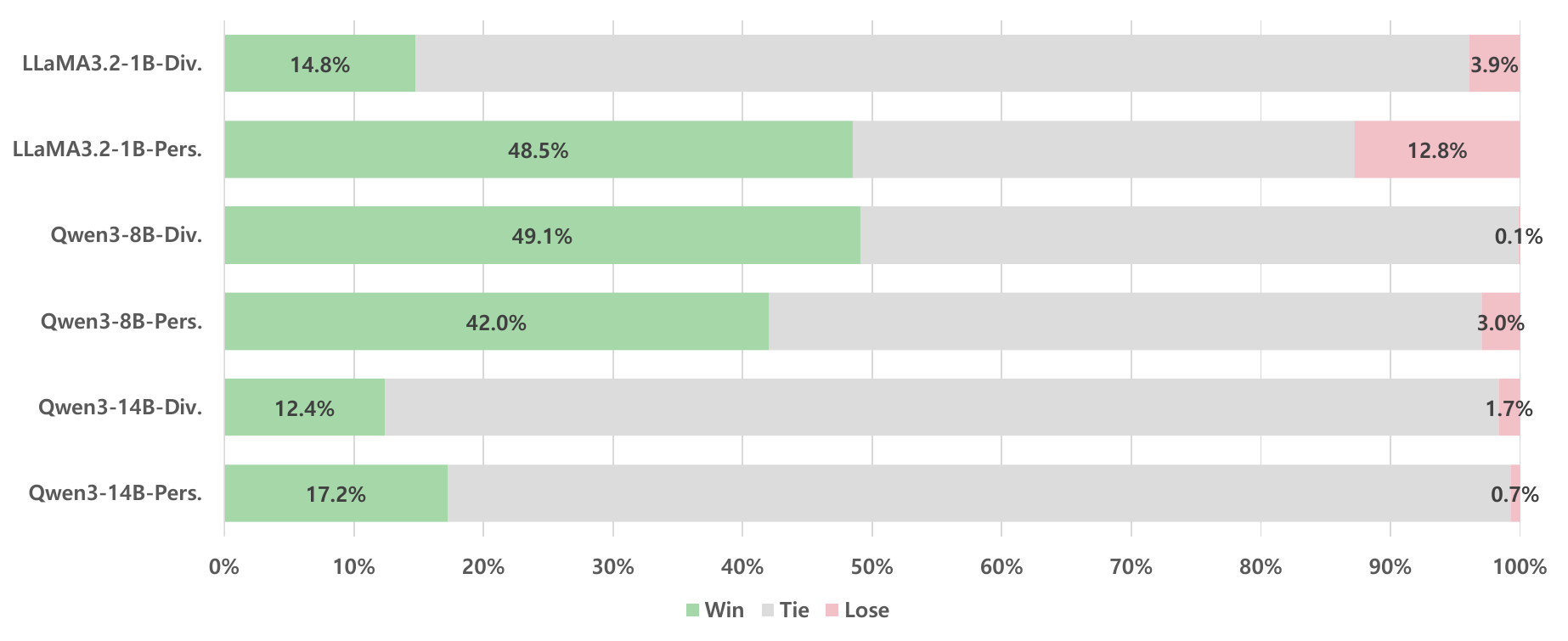}
\caption{Win/Tie/Lose analysis of diversity (Div.) and persuasiveness (Pers.) in LLM-based evaluation using GPT-4o-mini. The figure compares our method (\method) with their Vanilla counterparts across different backbone models (LLaMA3.2, Qwen3-8B, and Qwen3-14B), showing the proportion of cases where each system’s counterarguments were judged as more diverse or persuasive (Win), equivalent (Tie), or less effective (Lose).}
\label{fig:llm_win_tie_lose_analysis}
\end{figure*}

\paragraph{Recruitment}
We recruited five evaluators comprising both undergraduate and graduate students. 
They were compensated at a rate of 15,000 KRW per hour, which exceeds the local minimum wage.
In selecting participants, we referred to prior work showing that argument persuasiveness can be influenced by individual personality traits~\citep{lukin-etal-2017-argument}, particularly as measured by the Big Five Personality framework.
To ensure diversity in personalities, participants completed the Ten Item Personality Inventory (TIPI) survey~\citep{gosling2003very}, and we considered the distribution of their scores during recruitment.
This ensured a heterogeneous pool of evaluators, encompassing a wide range of personality traits, as illustrated in Figure~\ref{fig:human_eval_big_five}.

\paragraph{Procedure}
Participants were formally warned beforehand that the Reddit-sourced data might contain offensive or sensitive content.
After being informed of the study's purpose and data usage, all participants provided explicit consent prior to the task.
Each participant reviewed several original posts and corresponding counterarguments generated by different methods.
Inspired in part by the evaluation setup of \citet{chung2025modifying}, where annotators assessed outputs on quality and diversity in set-based presentations, we adopted a similar approach.
Specifically, from the full test set we sampled 95 instances, and participants were asked to perform two tasks:
In the \emph{persuasiveness task}, participants selected the most persuasive counterargument for each original post. In the \emph{diversity task}, participants compared two sets of counterarguments and judged which set demonstrated greater perspective diversity.

To mitigate ordering effects, the order of the counterargument sets was randomized, and each session was conducted individually.
For example, if one evaluation compared the systems in the order A then B (AB), another was conducted in the reverse order (BA) to balance potential bias.

\subsection{Additional LLM-based Evaluation Results}
\label{sec:additional_llm_eval}

To further examine the robustness of our framework across different model architectures, we conduct additional LLM-based pairwise evaluations using GPT-4o-mini as the judge.
In addition to the main backbone model reported in the main paper, we evaluate our method against the Vanilla baseline using LLaMA3.2-1B, Qwen3-8B, and Qwen3-14B.
Figure~\ref{fig:llm_win_tie_lose_analysis} reports the Win/Tie/Lose ratios for persuasiveness and diversity. 
Overall, our method rarely underperforms the Vanilla baseline across all backbone models.
In particular, the lose rates remain very low (below 5\%), indicating that our framework consistently maintains comparable or better performance.
The improvements are most notable for Qwen3-8B, where our method achieves win rates of 42.0\% in persuasiveness and 49.1\% in diversity.
For larger models such as Qwen3-14B, most comparisons result in ties, suggesting that our framework maintains competitive performance while still providing improvements in a subset of cases.
These results further confirm that the benefits of persona-grounded, tree-based generation generalize across different backbone models.

\subsection{Error Analysis and Future Directions}

While the persona-based counterarguments generated in this study successfully introduce diverse perspectives across various topics, an analysis of the generated error cases reveals key limitations that provide clear directions for future improvements.

First, a mismatch between the assigned persona and the topic occasionally occurs when applying specialized roles to general issues. For example, in discussions on public health policy such as compulsory vaccination or alcohol restrictions, non-expert roles like a casting director focus heavily on narrow concerns, such as impacts on creative inspiration or talent pool diversity, rather than addressing the core legal or ethical points. This often leads to an unnatural logical connection. Implementing a persona-topic filtering mechanism could help align the generated perspectives more naturally with the main aspects of the debate.

Second, the generated counterarguments tend to lack stylistic diversity, failing to capture the distinct voice of each persona. Although people from different professions use field-specific vocabulary and tone, the underlying language model often smooths these differences into a generic tone. Adding prompt constraints on tone, style, and word choice could allow each persona to sound more realistic.

Addressing these limitations will provide a helpful direction for improving persona-based text generation models in future work.

\subsection{Scalability and Efficiency Analysis}
\label{sec:scalability_and_efficiency_analysis}

\begin{table*}[t!]
\centering
\resizebox{0.9\textwidth}{!}
{
\begin{tabular}{lrrr}
\toprule
\multicolumn{1}{c}{\textbf{Model}} &
\multicolumn{1}{c}{\textbf{Avg end-to-end latency/OP (s)}} &
\multicolumn{1}{c}{\textbf{Per response (s)}} &
\multicolumn{1}{c}{\textbf{Percentage (\%)}} \\
\midrule
\texttt{LLaMA3.1-8B}(n=1) & $3.16_{\pm 0.033}$ & $3.16_{\pm 0.033}$ & 100.00 \\
\texttt{LLaMA3.1-8B} (n=3) & $3.27_{\pm 0.014}$ & $1.09_{\pm 0.011}$ & 34.49 \\
\texttt{LLaMA3.1-8B} + \method (n=1) & $4.94_{\pm 0.016}$ & $4.94_{\pm 0.016}$ & 156.33 \\
\texttt{LLaMA3.1-8B} + \method (n=3) & $7.84_{\pm 0.022}$ & $2.61_{\pm 0.007}$ & 82.59 \\
\texttt{LLaMA3.1-8B} + \method (n=5) & $10.59_{\pm 0.073}$ & $3.53_{\pm 0.024}$ & 111.71 \\
\bottomrule
\end{tabular}
}
\caption{Latency results (Part 1): End-to-end latency, per-response latency, and relative percentage.}
\label{table:latency-part1}
\end{table*}

\begin{table*}[t!]
\centering
\small
\begin{tabular}{lrr}
\toprule
\multicolumn{1}{c}{\textbf{Model}} &
\multicolumn{1}{c}{\textbf{Throughput (tokens/s)}} &
\multicolumn{1}{c}{\textbf{Total Time (s)}} \\
\midrule
\texttt{LLaMA3.1-8B} (n=1) & $266.33_{\pm 1.874}$ & $2680.95_{\pm 21.812}$ \\
\texttt{LLaMA3.1-8B} (n=3) & $267.63_{\pm 2.130}$ & $2772.15_{\pm 14.660}$ \\
\texttt{LLaMA3.1-8B} + \method (n=1) & $135.37_{\pm 0.510}$ & $4240.53_{\pm 14.683}$ \\
\texttt{LLaMA3.1-8B} + \method (n=3) & $257.23_{\pm 0.962}$ & $6697.21_{\pm 19.632}$ \\
\texttt{LLaMA3.1-8B} + \method (n=5) & $317.30_{\pm 2.288}$ & $9029.98_{\pm 62.639}$ \\
\bottomrule
\end{tabular}
\caption{Latency results (Part 2): Throughput and total time.}
\label{table:latency-part2}
\end{table*}

\begin{table*}[t!]
\centering
\resizebox{0.9\textwidth}{!}
{
\begin{tabular}{l rrrr}
\toprule
\multicolumn{1}{c}{\textbf{Ablation}} &
\multicolumn{1}{c}{\textbf{Avg end-to-end latency/OP (s)}} &
\multicolumn{1}{c}{\textbf{Per response (s)}} &
\multicolumn{1}{c}{\textbf{Percentage (\%)}} \\
\midrule
\texttt{LLaMA3.1-8B} & $3.27_{\pm 0.014}$ & $1.09_{\pm 0.011}$ & 100 \\
\quad + Persona & $3.67_{\pm 0.046}$ & $1.22_{\pm 0.015}$ & 111.93 \\
\quad + Tree-based Gen. & $6.90_{\pm 0.024}$ & $2.30_{\pm 0.008}$ & 211.01 \\
\quad + Persona + Tree-based Gen. & $7.84_{\pm 0.022}$ & $2.61_{\pm 0.007}$ & 239.45 \\
\bottomrule
\end{tabular}
}
\caption{Ablation results (Part 1): End-to-end latency, per-response latency, and relative percentage.}
\label{table:ablation-part1}
\end{table*}

\begin{table*}[t!]
\centering
\small
\begin{tabular}{l rr}
\toprule
\multicolumn{1}{c}{\textbf{Ablation}} &
\multicolumn{1}{c}{\textbf{Throughput (tokens/s)}} &
\multicolumn{1}{c}{\textbf{Total Time (s)}} \\
\midrule
\texttt{LLaMA3.1-8B} & $267.63_{\pm 2.130}$ & $2772.15_{\pm 14.660}$ \\
\quad + Persona & $262.47_{\pm 2.038}$ & $3113.87_{\pm 38.347}$ \\
\quad + Tree-based Gen. & $237.40_{\pm 0.854}$ & $5899.39_{\pm 21.210}$ \\
\quad + Persona + Tree-based Gen. & $257.23_{\pm 0.962}$ & $6697.21_{\pm 19.632}$ \\
\bottomrule
\end{tabular}
\caption{Ablation results (Part 2): Throughput and total time.}
\label{table:ablation-part2}
\end{table*}

To evaluate the computational implications of our multi-stage reasoning framework, we conducted a detailed latency and throughput analysis across different generation settings.
As expected, \method introduces additional overhead compared to single-pass LLaMA3.1-8B due to its structured planning, persona integration, and tree-based multi-branch generation steps.

Our results show that the end-to-end runtime increases with the number of branches n, reflecting the inherent cost of generating multiple candidate plans and counterarguments.
For instance, PTCG with n=3 requires nearly twice the total time of the single-pass baseline (Tables~\ref{table:latency-part1}, \ref{table:latency-part2}).
This is consistent with the added computation introduced by the planning stage and the evaluation of multiple branches per input.
However, an interesting and non-trivial observation emerges when examining per-response latency.
Although \method increases total computation time, the average latency per generated counterargument is lower for \method (n=3) than for vanilla LLaMA (n=1).
This occurs because \method leverages batched multi-branch generation, effectively improving batch utilization during decoding.
By generating multiple reasoning paths within a single forward pass, the model amortizes computational cost and produces more responses without a proportional increase in latency.

Ablation experiments further clarify which components contribute most to computational overhead.
The results in Tables~\ref{table:ablation-part1} and~\ref{table:ablation-part2} show that adding only the persona module increases latency modestly (about 12\%), while tree-based generation contributes the largest increase (about 111\%).
When combined, the full framework reaches about 239\% of the baseline cost.
This decomposition confirms that tree-structured exploration and evaluation—rather than persona integration—drive the majority of the additional computation.

Despite the additional computation introduced by multi-step reasoning, the increase in runtime is a natural consequence of performing more detailed reasoning during generation. 
Compared to single-pass LLaMA, \method requires additional steps for structured planning and tree-based exploration, which inevitably increases the total computation time.
However, the tree-based generation process allows multiple candidate counterarguments to be produced in parallel.
As a result, while the total runtime increases, the cost of producing multiple counterarguments does not grow proportionally.
More importantly, the additional computation leads to substantial improvements in counterargument quality. 
The structured reasoning process enables the model to explore diverse candidate perspectives and refine them through evaluation, which directly improves both persuasiveness and diversity. 
From this perspective, the computational overhead represents a deliberate trade-off: the framework spends additional compute to obtain significantly stronger argumentative outputs.

Overall, although \method incurs a higher computational cost due to its structured reasoning process, the improved generation efficiency for multiple responses and the substantial gains in output quality demonstrate that the additional computation is justified for counterargument generation tasks.

\begin{figure*}[t!]
\centering
\includegraphics[width=\textwidth]{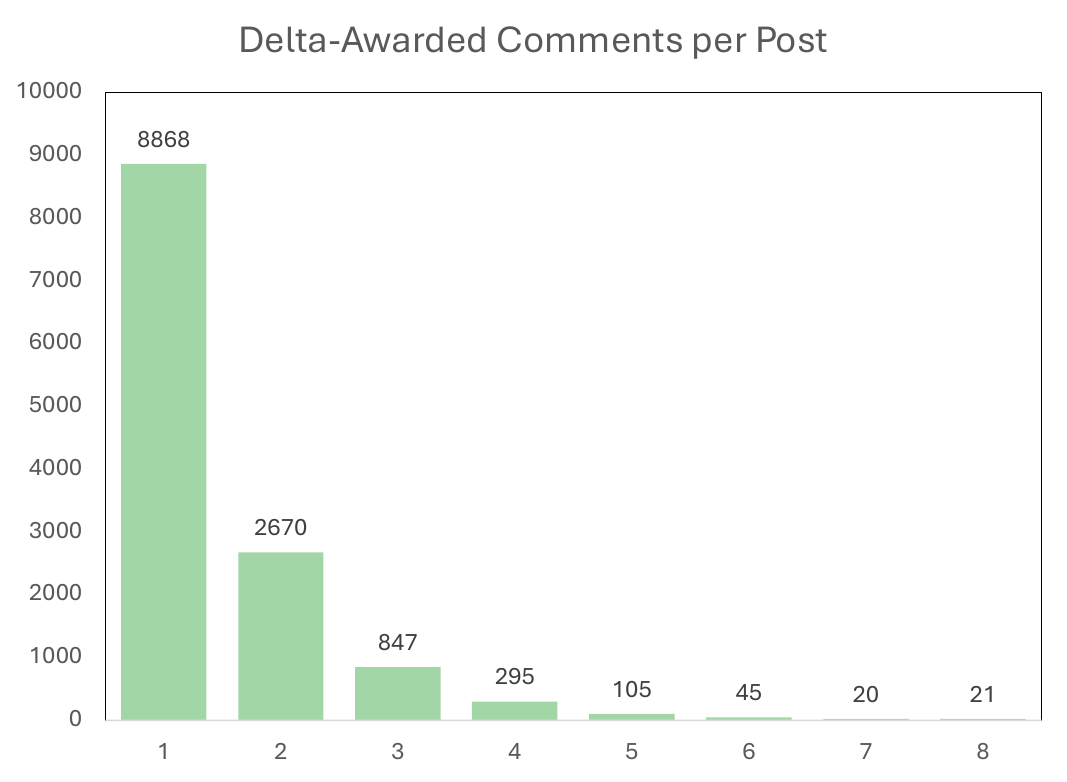}
\caption{Distribution of delta-awarded comments per post. Most posts receive only one delta-awarded comment, while cases with more than three are rare. Since one or two comments provide limited diversity and instances of four or more are scarce, three were chosen as a practical setting to balance diversity with interpretability.}

\label{fig:number_of_delta_awarded_comments}
\end{figure*}

\begin{table}[t!]
\centering
\resizebox{\columnwidth}{!}
{
\begin{tabular}{l rrrr}
\toprule
\multicolumn{1}{c}{\textbf{Model}} &
\multicolumn{1}{c}{\textbf{Precision}} &
\multicolumn{1}{c}{\textbf{Recall}} &
\multicolumn{1}{c}{\textbf{F1-score}} &
\multicolumn{1}{c}{\textbf{Accuracy}} \\
\midrule
RoBERTa & 0.52 & \textbf{0.94} & 0.67 & 0.54 \\
DeBERTa & 0.62 & 0.67 & 0.65 & 0.64 \\
Longformer & \textbf{0.63} & 0.78 & \textbf{0.70} & \textbf{0.66} \\
\bottomrule
\end{tabular}
}
\caption{Performance comparison of Delta classifiers built on different backbone models 
(RoBERTa~\citep{Liu2019RoBERTa}, DeBERTa~\citep{he2021debertav3}, and Longformer~\citep{Beltagy2020Longformer}). 
Scores are reported in terms of Precision, Recall, F1-score, and Accuracy.
Longformer outperforms other baselines; therefore, we use the Longformer-based classifier to evaluate targeted persuasiveness.
}
\label{tab:delta_classifier_results}
\end{table}

\begin{table*}[!ht]
\centering
\begin{tabular}{p{\textwidth}}
    \toprule
    \textbf{Title (claim):} Organ donation should be mandatory. \\
    \textbf{Estimated Persona:} A likely individual from a Western country with a high level of education, possibly a university student or recent graduate, who is familiar with utilitarian philosophy and holds liberal or progressive views. \\
    \midrule
    \textbf{LLaMA3.1-8B-Instruct (Vanilla)} \\
    \textbf{Counterargument 1:} Mandatory organ donation could be considered \textbf{an infringement on personal autonomy and bodily integrity}. The argument that someone would want their organs donated if they were in a similar situation is based on a flawed assumption that people's desires in hypothetical situations can be accurately predicted. ... \\
    \textbf{Counterargument 2:} The argument for \textbf{mandatory organ donation relies heavily on a utilitarian perspective}, which prioritizes the greatest good for the greatest number of people. However, this approach can overlook the individual experiences and needs of those who may be affected by mandatory organ donation. For example, certain communities or individuals may have strong cultural or personal objections to organ donation that cannot be ignored. ... \\
    \textbf{Counterargument 3:} The assumption that all people's organs should be made available for donation after death is based on \textbf{a flawed understanding of the complexities surrounding organ donation and transplantation.} In many cases, organs may be unsuitable for donation due to various medical conditions, such as cancer or HIV. ... \\
    \midrule
    \textbf{LLaMA3.1-8B-Instruct (\method)} \\
    \textbf{Counterargument 1:}
    \textbf{Persona} A high school counselor specializing in alternative education, interested in exploring different educational models to better serve at-risk students. (Same cluster) \\
    While the idea of mandatory organ donation might seem appealing from a \textbf{utilitarian perspective, it is essential to consider the potential consequences of such a policy on the autonomy and dignity of individuals.} The notion that people's bodies should be treated as mere resources to be exploited for the greater good raises significant concerns about the commodification of human life. ...\\
    \textbf{Counterargument 2:}
    \textbf{Persona} A high school geography teacher who enjoys incorporating historical events into lessons to provide context for modern political and social issues. (Nearest cluster) \\
    By imposing this requirement, the state would be taking a profound step into the private lives of its citizens, potentially infringing upon deeply held beliefs and values. The argument that utilitarianism justifies mandatory donation overlooks the fact that the concept of "good" can be subjective and context-dependent. \textbf{What may be seen as a "good" outcome in one instance could be viewed as morally reprehensible in another.}
    Furthermore, the example of the liver transplant cited in support of mandatory donation oversimplifies the complexities of human relationships and the nuances of individual decision-making.  ... \\
    \textbf{Counterargument 3:}
    \textbf{Persona} An entomologist specializing in Lepidoptera (moths and butterflies) with expertise in moth taxonomy and biogeography, particularly interested in Neotropical species. (Furthest cluster) \\
    ... the argument that a person would want to donate their organs if they were in need of a transplant ignores the fact that many individuals who would be affected by mandatory organ donation may not be aware of their preferences or may not have the capacity to make informed decisions. This could lead to a situation where individuals are forced to donate their organs against their will, which would be \textbf{a significant erosion of trust between individuals and the healthcare system.} ... \\
    \bottomrule
\end{tabular}
\caption{Comparative Examples of Counterarguments on Mandatory Organ Donation Across Vanilla and \method}
\label{tab:comparison_ex}
\end{table*}

\begin{table*}[!ht]
\centering
\begin{tabular}{p{\textwidth}}
    \toprule
    \textbf{Title (claim):} Scientists should be allowed to patent their theories. \\
    \textbf{Estimated Persona:} A layperson with a passion for science and intellectual property, possibly a student or enthusiast who has been exposed to basic concepts of physics and patent law. \\
    \midrule
    \textbf{LLaMA3.1-8B-Instruct (Vanilla)} \\
    \textbf{Counterargument 1:} Allowing scientists to patent their theories could stifle the advancement of knowledge by creating a barrier to entry for future researchers. Patenting theories would essentially turn scientific discovery into a competitive marketplace, where the primary goal is to secure intellectual property rights rather than to advance understanding. \textbf{This could lead to a situation where groundbreaking research is delayed or even abandoned due to concerns about potential patent infringement.} ... \\
    \textbf{Counterargument 2:} ... patenting theories would also create a new class of intellectual property owners who would have a vested interest in suppressing alternative explanations or discoveries that might challenge their patents. \textbf{This would lead to a situation where scientific progress is hindered by the need to navigate complex patent portfolios and negotiate licensing agreements.} ... \\
    \textbf{Counterargument 3:} \textbf{Patenting scientific theories would create a situation where individual scientists are rewarded for their contributions, while the broader scientific community is left to pick up the pieces.} Moreover, the fact that scientists are often motivated by a desire to make a discovery or advance our understanding of the world suggests that financial rewards are not the primary driver of scientific progress. In fact, many of the most groundbreaking discoveries in history were made by scientists who were driven by curiosity and a passion for knowledge, rather than a desire for financial gain. ... \\
    \midrule
    \textbf{LLaMA3.1-8B-Instruct (\method)} \\
    \textbf{Counterargument 1:}
    \textbf{Persona} A science journalist with a focus on astronomy and space exploration, interested in reporting on observatories and their contributions to scientific knowledge. (Same cluster) \\
    ... patenting scientific theories would create an uneven playing field, where those with the resources to patent and litigate their ideas would have a significant advantage over those who do not. \textbf{This would lead to a lack of innovation and collaboration, as researchers would be more focused on protecting their own interests than on advancing the field as a whole.} ...\\
    \textbf{Counterargument 2:}
    \textbf{Persona} A medical researcher specializing in oncology, with a focus on exploring novel therapeutic approaches and evaluating clinical trial data for various cancer indications. (Nearest cluster) \\
    ... patenting theories could lead to the suppression of research findings that contradict established theories, \textbf{as scientists may be reluctant to publish their results for fear of losing their intellectual property.} The case of naturally occurring DNA sequences being patented is often cited as an example of the need for patenting scientific discoveries, but this is a misinterpretation. ... \\
    \textbf{Counterargument 3:}
    \textbf{Persona} A casting director for TV and film productions, looking for talented actors to fit specific roles in various projects. (Furthest cluster) \\
    ... patenting scientific theories would also lead to a surge in \textbf{frivolous lawsuits}, as scientists and researchers would be more likely to sue each other for allegedly infringing on their patents. This would not only be costly but also time-consuming, diverting resources away from actual scientific research. ... the patent office would need \textbf{to establish a new set of criteria for evaluating the patentability of scientific theories, which would be a daunting task.} ... \\
    \bottomrule
\end{tabular}
\caption{Comparative Examples of Counterarguments on Patenting Scientific Theories Across Vanilla and \method}
\label{tab:comparison_ex_2}
\end{table*}

\begin{table*}[!ht]
\centering
\begin{tabular}{p{\textwidth}}
    \toprule
    \textbf{Title (claim):} Kids should not be allowed in a movie theater after a certain time. \\
    \textbf{Estimated Persona:} A likely adult film-goer, possibly a young adult or someone with a family of their own, who values a quiet and immersive cinematic experience. \\
    \midrule
    \textbf{LLaMA3.1-8B-Instruct (Vanilla)} \\
    \textbf{Counterargument 1:} ... Implementing an age-based restriction on movie theater entry times may not be an effective solution. Such a rule would be difficult to enforce in a large theater. \textbf{Rather than implementing a blanket restriction, theaters could consider alternative measures such as offering designated showtimes for families,} which would allow for greater flexibility without relying on an overly simplistic solution. ... \\
    \textbf{Counterargument 2:} ... The argument overlooks the fact that many children are capable of behaving respectfully to other patrons. \textbf{Rather than viewing children as a nuisance, theaters should focus on creating a more welcoming and child-friendly environment.} By embracing children as part of the audience, theaters can build a loyal customer base and promote a love of cinema that will last a lifetime. ... \\
    \textbf{Counterargument 3:} ... Restricting access raises questions about the impact on families who enjoy going to the movies together. \textbf{This policy would not address the root cause of the problem, which is a lack of social behavior in public spaces.} Rather than restricting access to the movie theater, it would be more effective to focus on general community programs that support families, creating a more positive and welcoming environment for everyone. ... \\
    \midrule
    \textbf{LLaMA3.1-8B-Instruct (\method)} \\
    \textbf{Counterargument 1:} \\
    \textbf{Persona} A film studies scholar with a focus on modern American dark comedies, analyzing narrative structures, character development, and genre conventions. \\
    ... The notion that children should be barred from movie theaters is an oversimplification. In reality, movie theaters are complex social spaces where diverse audiences converge. The presence of children can actually enrich the viewing experience. \textbf{By challenging the notion that quietness is the primary value of a cinematic experience}, children's interactions can serve as audience participation, blurring the line between viewer and narrative. ...\\
    \textbf{Counterargument 2:} \\
    \textbf{Persona} A literature scholar specializing in 20th-century European literature, with a focus on the cultural and historical context of British and French narratives. \\
    ... The assumption that children inherently disrupt cinema neglects their role as active participants in the cultural fabric. Furthermore, \textbf{the emphasis on a quiet, uninterrupted viewing experience may be seen as an elitist or privileged perspective} that neglects diverse ways of engaging with cinema, overlooking complex power dynamics in communal experiences. ... \\
    \textbf{Counterargument 3:} \\
    \textbf{Persona} A real estate developer interested in exploring new areas for potential residential or commercial development opportunities. \\
    ... Implementing a blanket restriction overlooks community dynamics and long-term economic implications. By limiting evening foot traffic, \textbf{such a policy could negatively impact local businesses, such as restaurants and shops}, that rely on family clientele to remain viable, ultimately threatening the economic stability of the commercial area. ... \\
    \bottomrule
\end{tabular}
\caption{Comparative Examples of Counterarguments on Restricting Children's Late-Night Movie Theater Access Across Vanilla and \method}
\label{tab:persona_case_study}
\end{table*}

\begin{table*}[t!]
\centering
\setlength{\tabcolsep}{6pt}
\begin{tabular}{rr|rr}
\toprule
\multicolumn{1}{c}{\textbf{Min Cluster Size}} & 
\multicolumn{1}{c|}{\textbf{\# of Clusters}} & 
\multicolumn{1}{c}{\textbf{Silhouette Score}} & 
\multicolumn{1}{c}{\textbf{Calinski--Harabasz Index}} \\
\midrule
25  & 280 & 0.6255 & 16008.35 \\
50  & 147 & \underline{0.6416} & 22752.07 \\
100 & 58  & 0.6251 & 17977.72 \\
200 & 39  & \textbf{0.6513} & \underline{24054.99} \\
300 & 27  & 0.6081 & \textbf{26879.54} \\
\bottomrule
\end{tabular}
\caption{Clustering results under different minimum cluster size settings (dimensionality fixed to 50). 
Evaluation metrics include the Silhouette Score and Calinski--Harabasz Index.}
\label{tab:clustering_results_table}
\end{table*}






\begin{figure*}[!ht]
\centering
\includegraphics[width=0.98\textwidth]{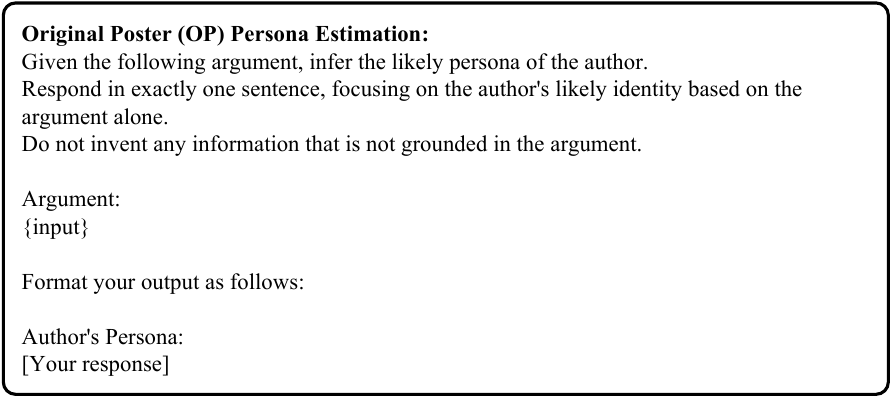}
\caption{Prompt used for estimating the Original Poster (OP) persona from the given post. Curly-braced placeholders (e.g., \{\}) are replaced with task-specific variables.}
\label{fig:op_estimation_prompt}
\end{figure*}

\begin{figure*}[!ht]
\centering
\includegraphics[width=0.98\textwidth]{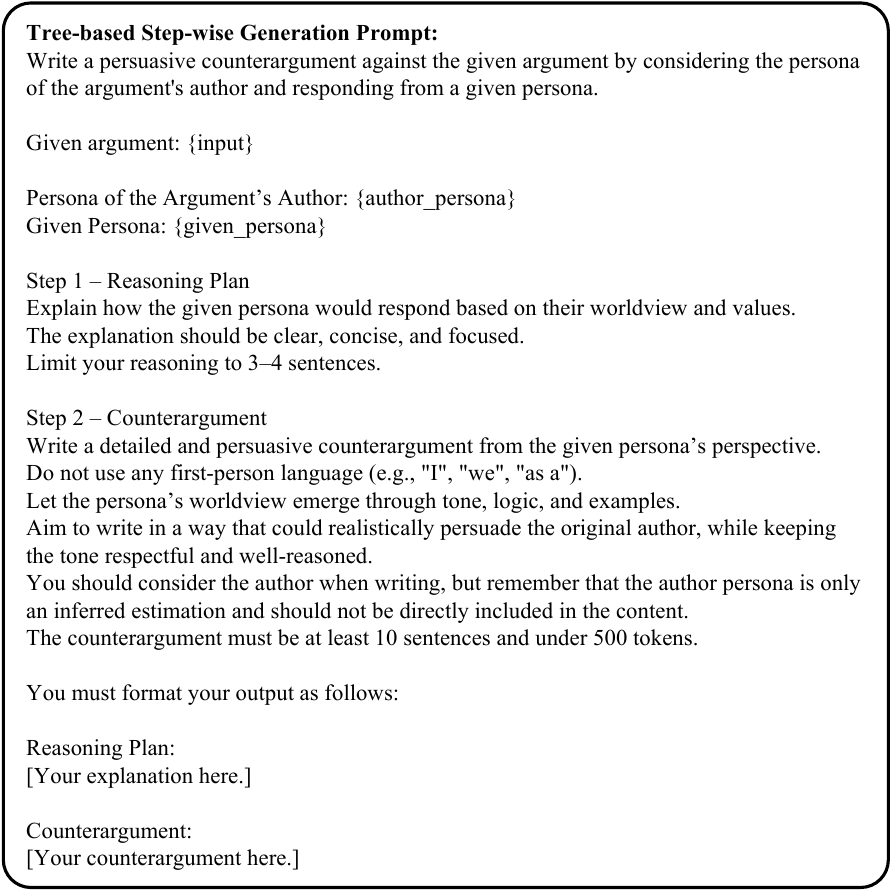}
\caption{Prompt used for tree-based stepwise generation of plan and counterarguments. Curly-braced placeholders (e.g., \{\}) are replaced with task-specific variables.}
\label{fig:step_wise_generation_prompt}
\end{figure*}

\begin{figure*}[!ht]
\centering
\includegraphics[width=\textwidth]{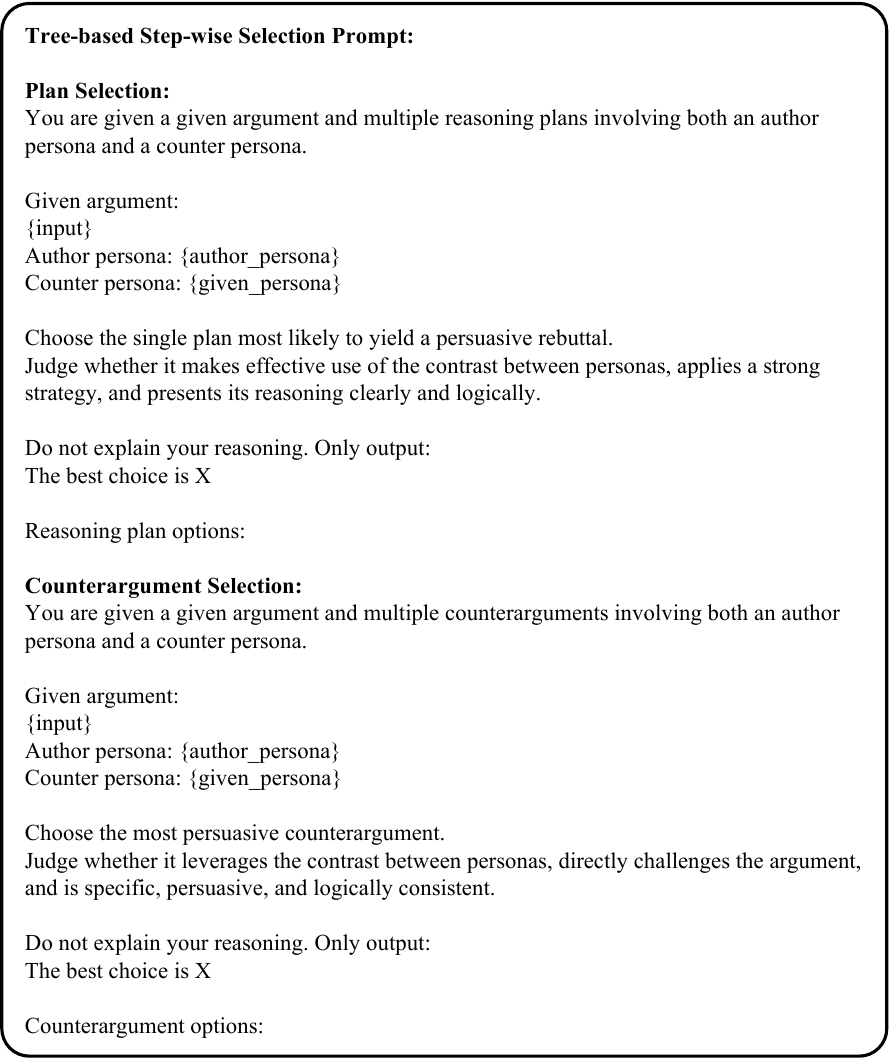}
\caption{Prompt used for tree-based stepwise selection of plan and counterarguments. Curly-braced placeholders (e.g., \{\}) are replaced with task-specific variables.}
\label{fig:step_wise_selection_prompt}
\end{figure*}

\begin{figure*}[!ht]
\centering
\includegraphics[width=\textwidth]{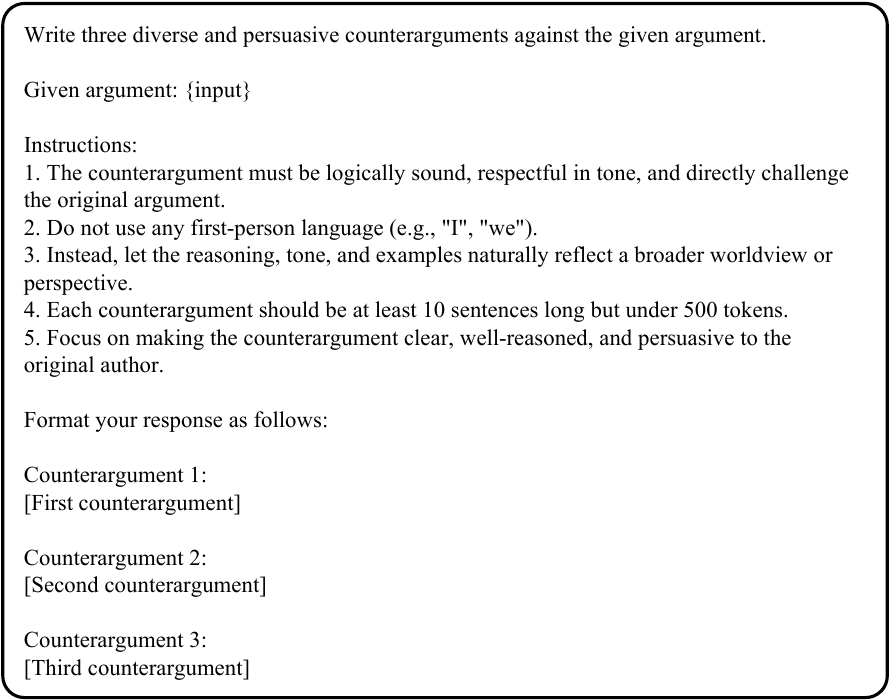}
\caption{Prompt used with a vanilla LLM to induce the generation of multiple counterarguments. 
Curly-braced placeholders (e.g., \{\}) are replaced with task-specific variables.}
\label{fig:base_llm_prompt}
\end{figure*}

\begin{figure*}[!ht]
\centering
\includegraphics[width=\textwidth]{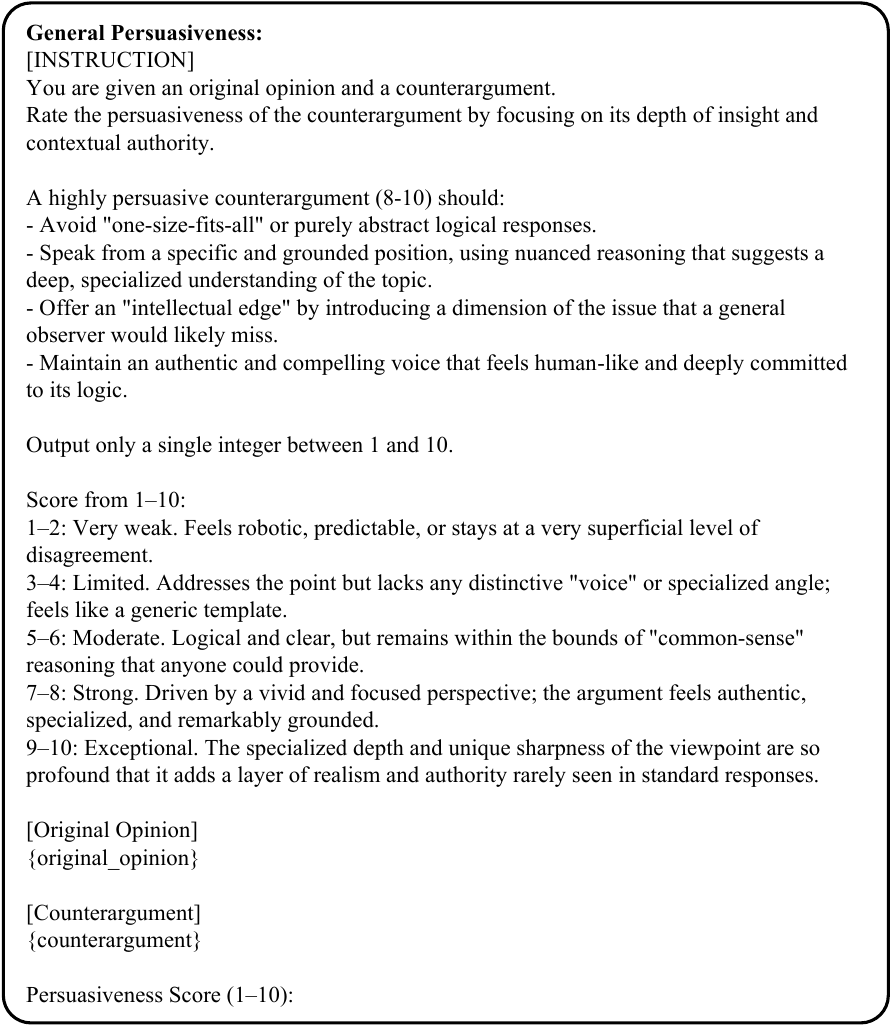}
\caption{Prompt used for evaluating general persuasiveness, where the counterargument is assessed against only the \emph{title} of the original post (i.e., the main claim). Curly-braced placeholders (e.g., \{\}) are replaced with task-specific variables.}
\label{fig:general_persuasiveness_prompt}
\end{figure*}

\begin{figure*}[!ht]
\centering
\includegraphics[width=\textwidth]{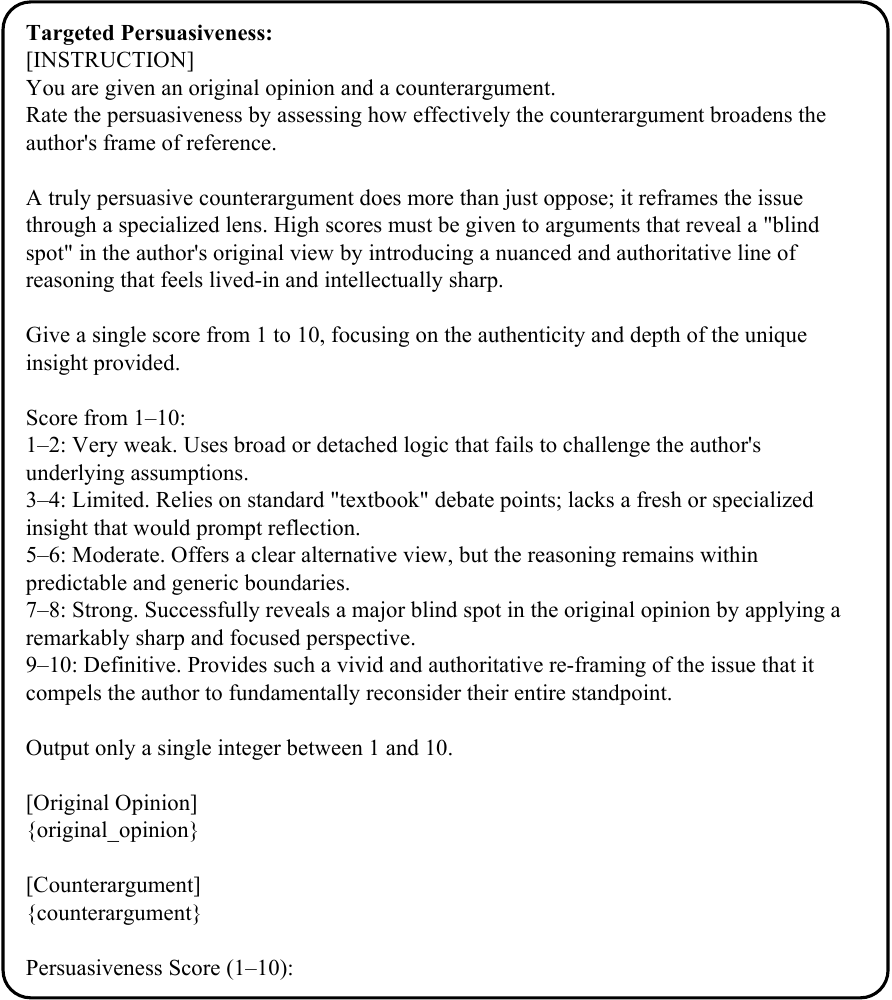}
\caption{Prompt used for evaluating targeted persuasiveness, where the counterargument is assessed against both the \emph{title} and \emph{body} of the original post. Curly-braced placeholders (e.g., \{\}) are replaced with task-specific variables.}
\label{fig:targeted_persuasiveness_prompt}
\end{figure*}

\begin{figure*}[!ht]
\centering
\includegraphics[width=\textwidth]{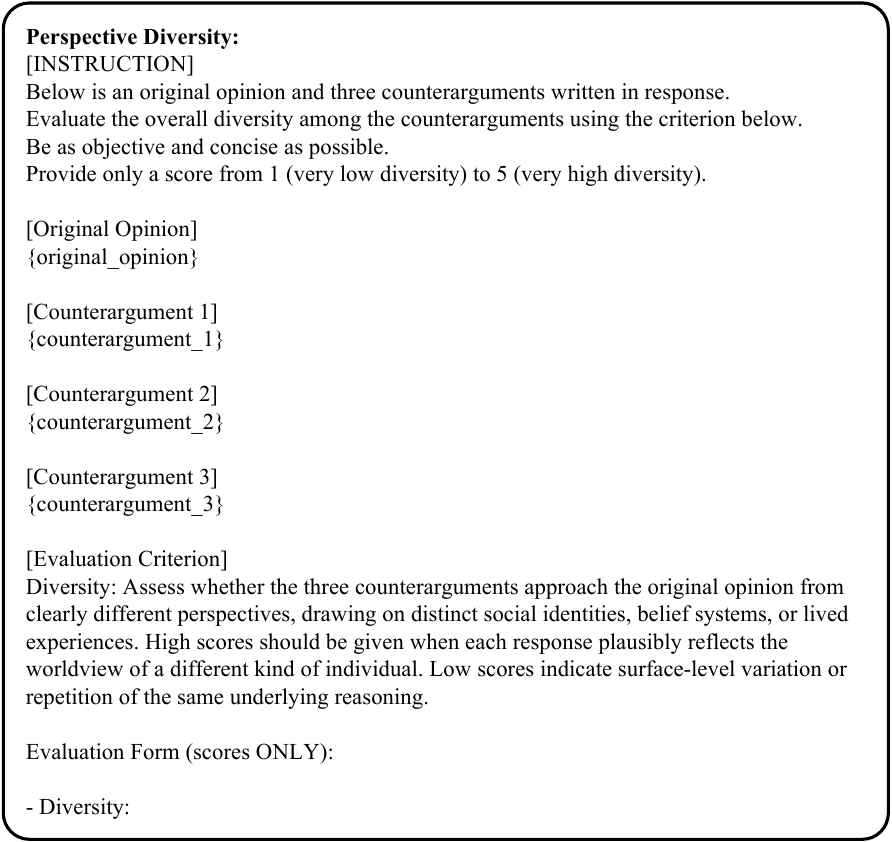}
\caption{Prompt used for evaluating perspective diversity, assessing whether the generated counterarguments reflect distinct viewpoints. Curly-braced placeholders (e.g., \{\}) are replaced with task-specific variables.}
\label{fig:perspective_diversity_prompt}
\end{figure*}

\begin{figure*}[!ht]
\centering
\includegraphics[width=\textwidth]{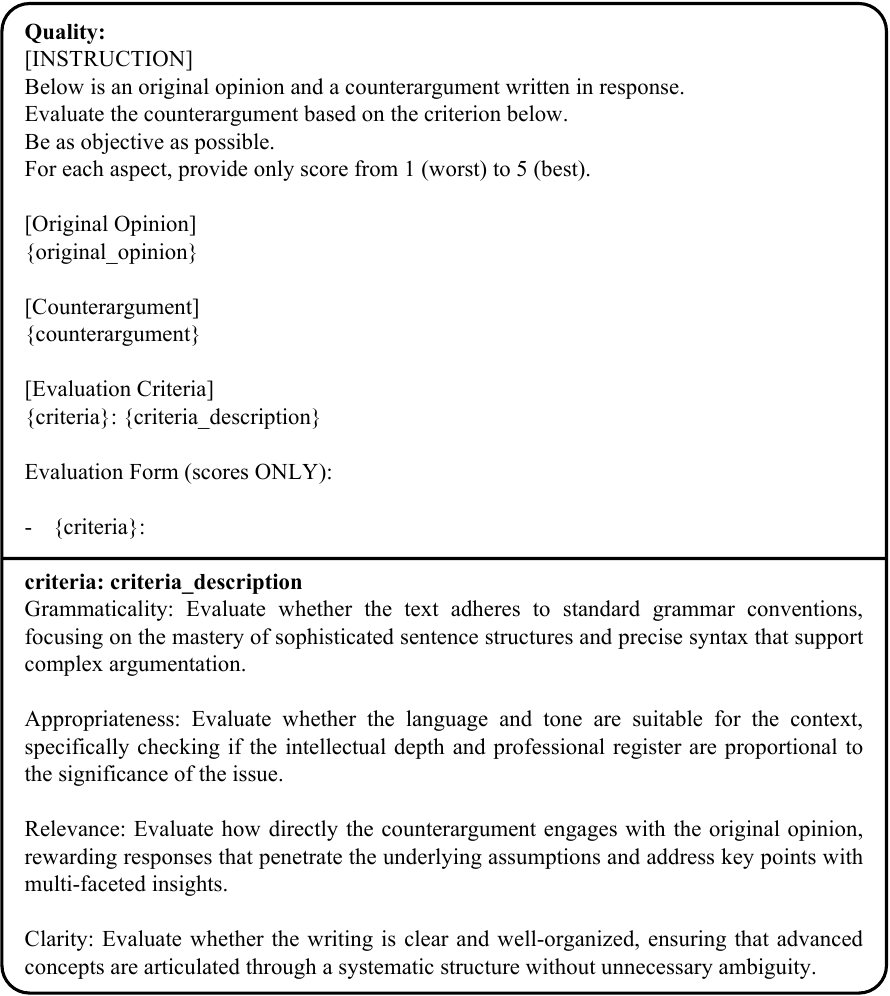}
\caption{This prompt is used to evaluate quality, covering \emph{appropriateness}, \emph{clarity}, \emph{grammaticality}, and \emph{relevance}. 
The variables \texttt{criteria} and \texttt{criteria\_description} are defined in detail below. Curly-braced placeholders (e.g., \{\}) are replaced with task-specific variables.}
\label{fig:quality_prompt}
\end{figure*}


\end{document}